%% file: main.tex
\documentclass[10pt,twocolumn,letterpaper]{article}

\usepackage{cvpr}              

\input{preamble}

\definecolor{cvprblue}{rgb}{0.21,0.49,0.74}
\usepackage[pagebackref,breaklinks,colorlinks,allcolors=cvprblue]{hyperref}

\theoremstyle{plain}
\newtheorem{theorem}{Theorem}[section]

\newtheorem{lemma}[theorem]{Lemma}
\newtheorem{corollary}[theorem]{Corollary}
\theoremstyle{definition}

\theoremstyle{remark}
\newtheorem{remark}[theorem]{Remark}

\def\paperID{56} 
\def\confName{CVPR}
\def\confYear{2026}

\title{OP-LoRA: The Blessing of Dimensionality with \underline{O}ver\underline{p}arameterized  \underline{Lo}w-\underline{R}ank \underline{A}daptation}

\author{
    Piotr Teterwak\textsuperscript{1} \quad
    Kate Saenko\textsuperscript{1} \quad
    Bryan A. Plummer\textsuperscript{1} \quad
    Ser-Nam Lim\textsuperscript{2}
    \\[0.7em]
    \textsuperscript{1}\textit{Boston University} \quad
    \textsuperscript{2}\textit{University of Central Florida}
}

\begin{document}

\twocolumn[{%
\maketitle

\begin{center}
    \newcommand{\teaserwidth}{\textwidth}
    \vspace{-7.5mm}
    \includegraphics[width=\teaserwidth]{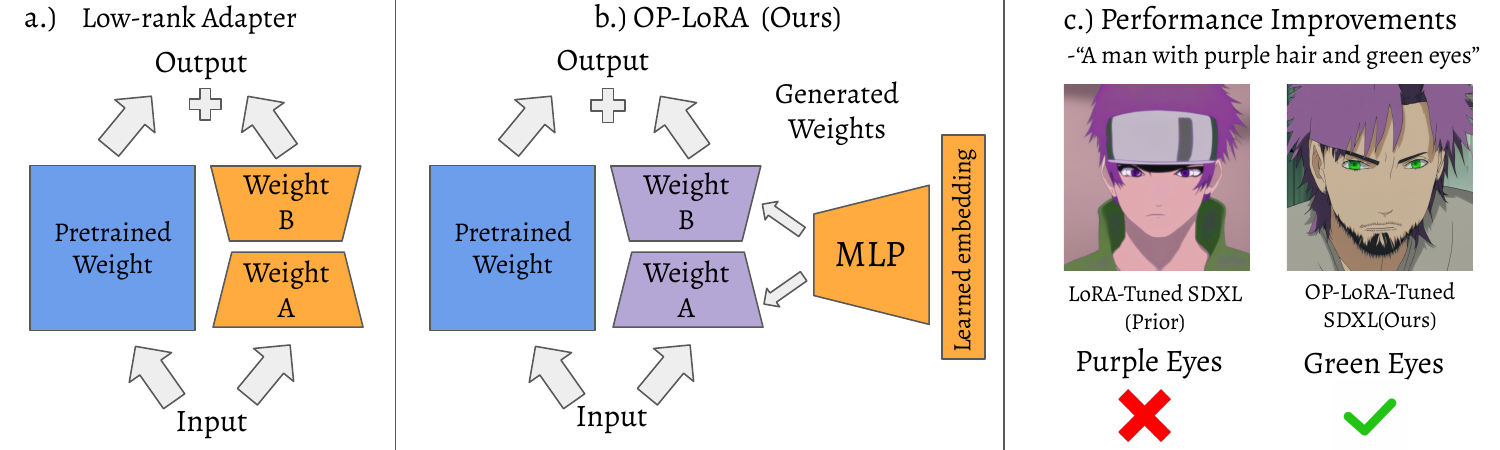}
    \captionof{figure}{\textbf{Overparameterizing low-rank adapters at training time improves performance.} (a) A standard Low-Rank Adapter learns two rank-reduced matrices (A and B) that are added to the frozen base weights. (b) Our proposed OP-LoRA  predicts the adapter weights from an MLP and a learned embedding. (c) Visual results of image generation on Stable Diffusion XL, showing qualitative improvements in generated images using our method compared to standard LoRA, in addition to being more faithful to the text prompt. (d) Performance in vision-language tasks, with OP-LoRA showing accuracy gains over LoRA across training epochs.  } 
	\label{fig:teaser}
\end{center}
}]

\input{sec/1_intro}
\input{sec/2_related_works}
\input{sec/4_Method}

\input{sec/6_Results}

\input{sec/8_conclusion}

{
    \small
    \bibliographystyle{ieeenat_fullname}
    \bibliography{main}
}

 \input{sec/X_suppl}

\end{document}

%% file: preamble.tex
\usepackage{amsmath}
\usepackage{booktabs}
\usepackage{makecell}
\usepackage{arydshln}
\usepackage{amsthm}
\usepackage{lmodern}

\newcommand{\W}{{\mathbf W}}

\def\be{\begin{equation}}
\def\ee{\end{equation}}
\def\beas{\begin{eqnarray*}}
\def\eeas{\end{eqnarray*}}
\def\bea{\begin{eqnarray}}
\def\eea{\end{eqnarray}}
\newcommand{\h}{{\mathbf h}}

%% file: sec/1_intro.tex
\begin{abstract}
Low-rank adapters (LoRA) enable  finetuning of large models with only a small number of parameters. However, they often suffer from an ill-conditioned loss landscape, leading to difficult optimization.   Prior work addresses these challenges by aligning adapter updates with full finetuning gradients via custom optimizers, but these methods lack the flexibility to accommodate new adapter architectures and are computationally expensive. We instead introduce OP-LoRA, a novel method which replaces each LoRA adapter with weights predicted by an extra MLP, which is discarded after training.  This  temporarily allows additional parameters during training to improve optimization, yet requires less wall time than custom optimizers and zero extra cost at inference time because the MLP is discarded. Crucially, extending OP-LoRA to other adapters is as simple as modifying the size of the prediction head for each new adapter type.  We show that OP-LoRA  allows the optimization to adaptively increase or decrease step size, improving performance and decreasing sensitivity to learning rate. On both small and large-scale LoRA tuning tasks, we observe consistent performance gains of OP-LoRA relative to LoRA and its variants. We achieve especially notable improvements in image generation, with OP-LoRA CMMD scores improving by up to 15 points relative to LoRA. This allows OP-LoRA to achieve the performance of LoRA with half of the inference parameters. 
\end{abstract}

\section{Introduction}
\label{sec:intro}

Finetuning large foundation models for specific tasks can provide significant performance gains but is computationally intensive, with risks of catastrophic forgetting~\citep{ruiz2023hyperdreambooth,cho2021unifying,biderman2024lora}. Methods utilizing low-rank adapters (LoRA)~\citep{hu2021lora,hayou2024lora+,zhang2023adalora,liu2024dora,nikdan2024rosa,meng2024pissa} address these challenges by modifying the model in rank-constrained ways (\cref{fig:teaser}a). Beyond reducing finetuning cost, this low-rank structure is now crucial for scalable inference, as systems such as S-LoRA~\citep{sheng2023slora} show that thousands of task-specific LoRA adapters can be served concurrently on a single base model with only small overhead. However, low-rank adapters can make optimization harder by creating uneven curvature in the loss landscape (Section \ref{sec:prelim}). Even when optimized with AdamW~\citep{loshchilov2019decoupled,kingma2015adam}, which preconditions gradients, poorly conditioned loss landscapes can still pose a problem~\citep{das2024towards}.  These issues manifest during LoRA training as high sensitivity to learning rates, as shown by~\citet{biderman2024lora} and confirmed in Section \ref{sec:mnist_lora}. 

The most successful attempts to address this issue are custom optimizers such as LoRA-Pro~\citep{anonymous2025lorapro} and ScaledAdamW~\citep{zhang2024riemannian}, which aim to align the LoRA update with that of full finetuning. However, in implementation, they are  complex and difficult to extend to new LoRA variants (see Limitations of~\citet{anonymous2025lorapro}). For example, adapting these optimizers to DoRA~\citep{liu2024dora} is non-trivial due to weight normalization, which complicates the projection of full finetuning gradients. They also tend to be more expensive to run than standard optimizers, requiring matrix inversions and expensive optimizations. In our testing using authors' code, wall time of ScaledAdamW is 15\% longer than that of OP-LoRA, and LoRA-Pro is up to 14x longer (Section \ref{sec:analysis}) when finetuning LLaMA on Commonsense Reasoning tasks. These limitations highlight the need for alternative optimization strategies that are both effective and architecture-agnostic.


Instead of relying on specialized optimizers, we propose a fundamentally different  and more flexible approach, which we call OP-LoRA (Overparameterized Low-Rank Adaptation). OP-LoRA uses a small MLP as a hypernetwork~\citep{ha2016hypernetworks} to predict the low-rank adapter matrices at \textbf{train time only} (Figure~\ref{fig:teaser}b).  In contrast to  other hypernetworks~\citep{ruiz2023hyperdreambooth,ortiz2024hyperloader}, we do not condition on the input sample in any way.  This allows us to discard the MLP at inference time, making inference and storage costs equal to that of standard LoRA at the same rank. Such low-rank inference footprints are particularly important in modern deployments, where systems like S-LoRA~\citep{sheng2023slora} rely on the small size of LoRA adapters to serve thousands of customized models on shared hardware. Interestingly, OP-LoRA preserves the  representational capacity of standard LoRA: any adapter produced by the MLP can be expressed exactly by conventional LoRA parameters. Yet, despite having no additional expressive power, OP-LoRA achieves improved performance. We show that this “blessing of dimensionality’’ arises from its enhanced ability to navigate complex loss landscapes through an acceleration mechanism (Section~\ref{sec:opt_benefits}).

Integration of OP-LoRA takes only a few lines of code and generalizes to any LoRA variant. For example, our OP-DoRA extension of DoRA simply adds a second MLP head to predict its extra adapter weights, something which would be difficult to do with custom optimizers. OP-LoRA outperforms standard variants by 1-6\% on natural language tasks  and up to 15 CMMD points on image generation tasks (Section \ref{sec:experiments}).

We  summarize our contributions as follows:

\begin{itemize}[itemsep=1pt, parsep=0pt, topsep=1pt, partopsep=0pt]
\item We introduce OP-LoRA, a novel yet easy-to-implement reparameterization of LoRA that uses an MLP to predict adapter weights instead of learning them directly. After training, the MLP is discarded so zero additional storage or inference costs are incurred.  
\item We show that OP-LoRA navigates loss landscapes better than standard LoRA due to a built-in acceleration mechanism (Section \ref{sec:opt_benefits}). 
\item We empirically validate OP-LoRA on a large range of tasks including image and text generation and show  consistent performance gains on both, and a large  improvement in adapting Stable Diffusion relative to standard LoRA (Section \ref{sec:experiments}). 
\end{itemize}

More generally, we believe that train-time over-parameterization represents a promising yet underexplored paradigm in model training, and we hope that our work will catalyze further work.

%% file: sec/2_related_works.tex
\section{Related Work}
\noindent\textbf{Low-rank finetuning:} Low-rank finetuning, specifically with LoRA (Low-Rank Adaptation)~\citep{huh2021low}, has emerged as a powerful approach for adapting pre-trained models with minimal additional parameters. A number of follow ups have emerged, improving performance.  AdaLoRA~\citep{zhang2023adalora} prunes weights during training.  DoRA~\citep{liu2024dora}  adds a magnitude scaling vector to the updated matrix.  RoSA~\citep{nikdan2024rosa} adds a sparse weight update to the low-rank update, but requires a full-finetuning pass to compute the weight mask. OLoRA~\citep{buyukakyuz2024olora} and PiSSA~\citep{meng2024pissa}  ease optimization by initializing LoRA orthogonally. However, they remove important pre-trained components from the frozen base weights. Another approach is to make the LoRA optimization trajectory similar to that of full finetuning: LoRA-GA~\citep{wanglora} initializes LoRA parameters to an SVD approximation of the full-finetuning gradient, while LoRA-Pro~\citep{anonymous2025lorapro} and ScaledAdamW~\citep{zhang2024riemannian} project full-tuning gradients onto the LoRA subspace. Both have computational overhead resulting in extended training times: LoRA-Pro requires expensive computations in gradient projection while LoRA-GA requires a full-finetuning pass similar to RoSA. Deep LoRA~\citep{yaras2024compressible}  learns an over-parameterized LoRA first before compressing it, leveraging the over-parameterization for improved training. However, the compression is an expensive process, and therefore impractical at scale. \\
\noindent\textbf{Reparameterization  with  hypernetworks:}~\citet{ha2016hypernetworks}  generate weights of an LSTM and CNN from a neural network, introducing the concept of HyperNetworks. However, their focus is relaxing weight sharing in LSTMs and reducing parameter count in convolutional networks for image classification. In contrast, we leverage over-parameterization for improved performance. HyperDreamBooth~\citep{ruiz2023hyperdreambooth} generates initializations for LoRA parameters from an input image. Instead, we learn our parameter-generating MLP with a learned parameter vector as input. HyperLoader~\citep{ortiz2024hyperloader} uses a hypernetwork to generate adapters, but shares parameters between layers and tasks. We show that this shared structure severely reduces performance in Table \ref{tab:commonsense}.\\
\noindent\textbf{Convergence Properties of Neural Networks:} Understanding the convergence behavior of neural networks has been a subject of significant research interest~\citep{du2019gradient,nguyen2017loss}.~\citet{du2019gradient} find that gradient descent can find global optima in ResNets.~\citet{huang2020understanding} find that optimizers are biased to flat minima in overparameterized models,  and coin the term the ``blessing of dimensionality''. Most relevant to OP-LoRA, though, is~\citet{arora2018optimization}'s work showing that stacking linear layers can function as an implicit acceleration mechanism in gradient descent. \\ 

%% file: sec/4_Method.tex
\section{Methodology: Overparameterized LoRA (OP-LoRA)}
\label{sec:prelim}

Low-Rank Adaptation (LoRA) has become a popular strategy for finetuning large models, allowing adaptation to new tasks by learning a low-rank matrix factorization of weight updates.  With LoRA, finetuning a model layer’s weight matrix \( W_0 \in \mathbb{R}^{d \times d} \) is achieved by learning an additive low-rank update \( \Delta W \) such that the adapted weights \( W \) are given by:
\begin{equation*}
\label{eq:lora}
W = W_0 + \Delta W = W_0 + BA,
\end{equation*}
where \( A \in \mathbb{R}^{r \times d} \) and \( B \in \mathbb{R}^{d \times r} \) are learned low-rank matrices, with \( r \ll d \), reducing the number of parameters to learn.  However, LoRA introduces challenges during optimization. While the original parameter space has curvature defined by the Hessian \( H_W \), the LoRA $A$ matrix has the transformed Hessian, represented as a composition of functional operators:
\begin{equation*}
\label{eq:hessian}
H_A = B^\top \circ\ H_W\circ B.
\end{equation*}
This transformation affects the condition number of the optimization problem in the \(A\)-space. Even if the original Hessian \( H_W \) is well-conditioned and symmetric positive definite (a reasonable assumption near local minima), the reparameterized Hessian can become ill-conditioned depending on the singular values of \( B \). In particular, the condition number \( \kappa(H_A) \) satisfies:
\begin{equation*}
 \frac{\kappa(B)^2}{\kappa(H_W)} 
\;\leq\;
\kappa(H_A)
\;\leq\;
\kappa(B)^2 \cdot \kappa(H_W)
\end{equation*}
where \( \kappa(\cdot) \) denotes the spectral condition number. A higher condition number indicates a greater ratio between the largest and smallest curvatures of the loss landscape.  In practice, higher condition numbers lead to slower convergence and greater sensitivity to learning rates. These bounds imply that even if \( H_W \) is well conditioned, poor conditioning in \( B \) alone can lead to difficulty in optimizing \( A\), since they drive up both upper and lower bounds. ~\citet{biderman2024lora} observe that LoRA is sensitive to learning rate, which we confirm in Section \ref{sec:mnist_lora}. The full derivation of this result,  is provided in the Supplementary Materials. In Section \ref{sec:opt_benefits} and \ref{sec:mnist_lora}, we  show OP-LoRA can dynamically adjust step size and perform an adaptive line search, overcoming the optimization difficulties of standard LoRA. In Section \ref{subsec:structural} of the Supplementary Materials, we analyze gradient properties of the Hessian of a trained model, and find evidence for OP-LoRA being less sensitive to poor conditioning than LoRA.

\begin{figure*}[t]
    \centering
    \includegraphics[width=\textwidth]{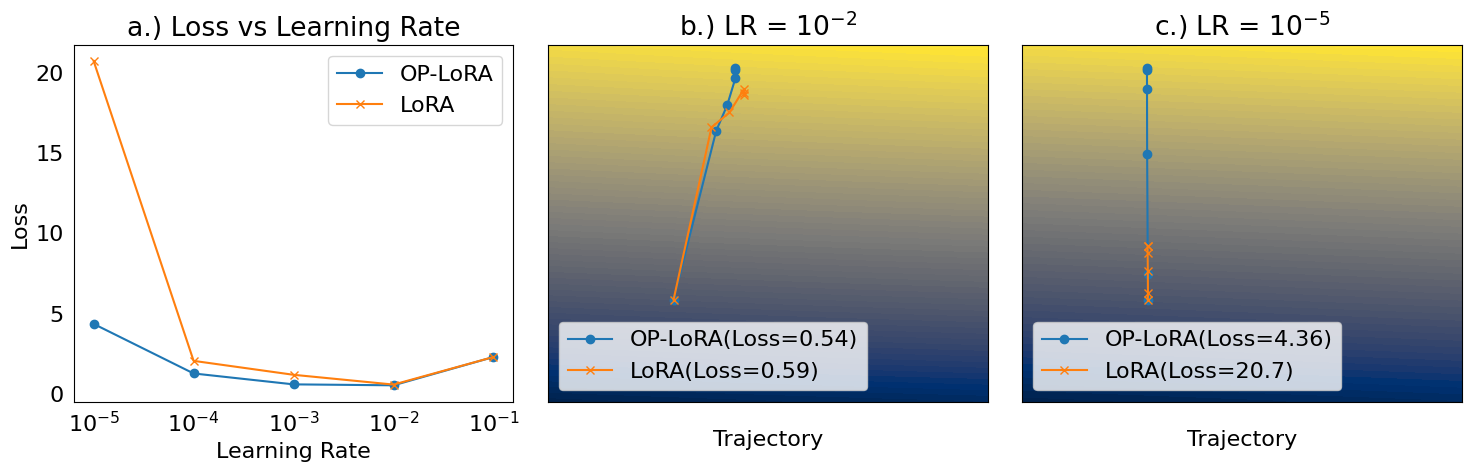}
    \caption{ \textbf{Optimization behavior of LoRA and OP-LoRA for Rotated MNIST classification}: \textbf{a.)} Training loss  achieved by LoRA and OP-LoRA as a function of the learning rate, showing that OP-LoRA  attains lower loss across a wide range of learning rates and remains robust even at suboptimal step sizes. \textbf{(b)} and \textbf{(c)} Parameter‐update trajectories overlaid on the training loss surface at (b) the optimal learning rate and (c) a low learning rate. In both cases, OP-LoRA  descends more directly towards lower loss (yellow contours) than LoRA. } 
    \label{fig:mnist_trajectories}
\end{figure*}

\subsection{Predicting LoRA Weights} In order to avoid the optimization difficulties of training LoRA discussed in Section \ref{sec:prelim},  we avoid directly optimizing \( A \) and \( B \) by introducing a two-layer MLP which takes as input \( z \) and predicts the entries of \( A \) and \( B \). Because the additional parameters are discarded after training, they \textbf{do not increase finetuning capacity}. Instead, this is a way of reshaping the optimization landscape, making it easier to optimize.  This is an important distinction with other methods of adding parameters to the finetuning procedure like increasing LoRA rank. Concretely, we generate \( A \) and \( B \)  as flattened matrices via:
\begin{equation*}
\begin{aligned}
    \begin{pmatrix} A \\ B \end{pmatrix} &= \W_2 (\text{ReLU}(\W_1 z + c_1)) + c_2
\end{aligned}
\end{equation*}
where \( z \) is the learned input vector to the MLP,  \( \W \) and \( c \) correspond to learned weights and biases, and  \( A \in \mathbb{R}^{r \times d} \) and \( B \in \mathbb{R}^{d \times r} \) are the generated  matrices.  

Once finetuning is complete, the MLP can be discarded, retaining only the low-rank matrices \( A \) and \( B \) for inference and storage. Furthermore, as in standard LoRA, \( A \) and \( B \) can be merged with the pre-trained model's weights by adding \( \Delta W = BA \) to the relevant layer weights (Equation \ref{eq:lora}). Although the MLP is compact in depth, it predicts a high-dimensional output, the size of the LoRA parameters, increasing its parameter count. For instance, an MLP with a hidden dimension of 32 scales the number of trainable parameters by approximately 32. This makes OP-LoRA particularly advantageous in settings where inference resources are constrained, but sufficient memory is available during training. Further, because the MLP is small relative to a typical base model, wall-time penalties are not large. For more  details on computational cost, see Section \ref{sec:analysis}.

\subsection{Optimization benefits of OP-LoRA }
\label{sec:opt_benefits}
\citet{arora2018optimization} prove that increasing depth by replacing linear layers with products of matrices, which has the same expressive power as a single matrix due to the linear nature of the transformation, leads to faster convergence.  Although we focus on re-parameterizing with an MLP instead of increasing depth, we employ the same theoretical framework to examine OP-LoRA’s enhanced training dynamics below. While we consider the linear case here for clarity, we extend the analysis to the  MLP case in the Supplementary Materials. 

Consider the OP-LoRA reformulation of parameter vector $v$ with a two-layer MLP, defined as  $v =  \W_2 (\text{ReLU}(\W_1 z +c_1)) + c_2$. We can then assign vector $h = \text{ReLU}(\W_1 z + c_1) $, and for clarity of derivation, we treat $h$ as a free parameter vector; this corresponds to only updating the bias in the first layer of the MLP. We also merge bias parameters into the parameter matrices $\W$ for ease of notation, assuming a constant value is appended to $z$ and $h$. This leaves us with the simpler reparameterization form $v = \W_2 h$,  where $v \in \mathbb{R}^p,\W_2 \in \mathbb{R}^{p \times k},\ h \in \mathbb{R}^k$ and $p$ is the size of generated parameter vector $v$ and $k$ is the hidden dimension of the MLP re-parameterization. 

\begin{figure*}[t!]
    \centering
    \includegraphics[width=\textwidth]{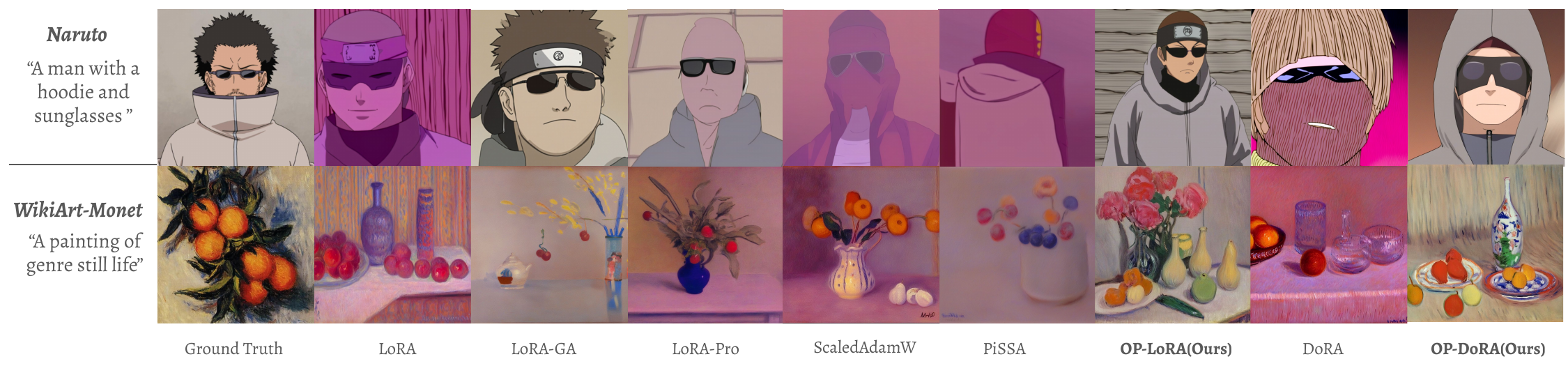}
    \caption{ \textbf{Comparison of generated images across different low-rank finetuning methods of SD-XL for two datasets:}  Naruto (upper) and Claude Monet-style painting (lower). Each row shows outputs from a specific model configuration based on ground truth captions.  For the Naruto prompt, OP-LoRA and OP-DoRA effectively capture the presence of a hoodie, and generally generate higher fidelity images. For the Monet-style paintings,  OP-LoRA and OP-Dora offer more realistic scenes.  } 
    \label{fig:qualitative}
\end{figure*}

The update rule for $\W_2$ and $h$ with learning rate $\eta$ is given by:
\begin{equation*}
\W_2^{(t+1)} = \W_2^{(t)} - \eta \nabla_{\W_2}, \quad 
h^{(t+1)} = h^{(t)} - \eta \nabla_{h}
\end{equation*}
Then, the parameter vector  $v$  becomes:
\begin{equation*}
v^{(t+1)} = \W_2^{(t+1)} h^{(t+1)} = \left( \W_2^{(t)} - \eta \nabla_{\W_2} \right) \left( h^{(t)} - \eta \nabla_{h} \right)
\end{equation*}
Expanding the product, we have:
\begin{equation*}
v^{(t+1)} = \W_2^{(t)} \h^{(t)} 
- \eta  \nabla_{\W_2} h^{(t)}
- \eta \W_2^{(t)} \nabla_{h} 
+ \eta^2 \nabla_{\W_2} \nabla_{h}
\end{equation*}
Following~\citep{arora2018optimization,zhang2024riemannian}, we  ignore the higher order term since the learning rate $\eta$ is assumed to be small and therefore the term shrinks to 0. By definition, $v = \W_2h$, so:
\begin{equation*}
v^{(t+1)} = v^{(t)} 
- \eta  \nabla_{\W_2} h^{(t)}
- \eta \W_2^{(t)} \nabla_{h} 
\end{equation*}
By substituting chain-rule expansions of $\nabla_{\W_2}$ and $\nabla_{h}$, the new value for $v$ becomes:
\begin{equation*}
\nabla_{\W_2} = \nabla_{v}  h^T, \quad 
\nabla_{h} =  \W_2^T  \nabla_{v}.
\end{equation*}
\begin{equation*}    
v^{(t+1)} 
= v^{(t)} 
- \underbrace{\eta \,\|h^{(t)} \|^2 \nabla_{v^{(t)}}}_{\text{trainable learning rate}} 
- \underbrace{\eta \,\W^{(t)}_2 \bigl((\W^{(t)}_2)^T \nabla_{v}\bigr)}_{\text{adaptive line search}}
\end{equation*}

This reveals two key properties of the optimization trajectory under OP-LoRA. First, OP-LoRA introduces a dynamic learning rate scaling factor, \(\|h^{(t)}\|^2\). 
From the update rule \(\nabla_{h} = (\W_2^{(t)})^T \nabla_{v}\), we see that 
\(\|h^{(t)}\|^2\) grows when the gradient descent direction is positively aligned with \(\W_2^{(t)}\) 
and shrinks otherwise. In the special case where \(k=1\), 
the vector \(h\) becomes a scaling factor for the single column vector \(\W_2\) which represents the direction of parameter vector $v$. Consequently, if the optimizer overshoots a minimum, the sign of the gradient update for $h$ flips, and 
the effective learning rate decreases, but if consecutive updates align, 
it increases. In the more general $k>1$ setting, there are $k$ such scalar factors (one for each column in $\W_2^{(t)}$) and each can increase or decrease independently, and the overall learning rate becomes scaled by  \(\|h^{(t)}\|^2\) .  We refer to \(\|h^{(t)}\|^2\) as the 
\textbf{\emph{trainable learning rate}}.

Second,  OP-LoRA adds an extra update term 
\(\W_2^{(t)} \bigl((\W_2^{(t)})^T \nabla_{v}\bigr)\),
which shifts parameters along the already learned directions \(\W_2^{(t)}\), 
in proportion to the gradient’s projection onto those directions. 
When $k=1$, this can be seen as a gradient step in a line-search, along the direction of the current state of $v$, and naturally biases the updates toward directions already taken. However,  if  $\nabla_{v}$ suddenly changes to a new and orthogonal direction, the final term immediately vanishes, causing the effective step size to shrink right away and the update  to suddenly shift towards the current gradient. For $k>1$, $v$ becomes a weighted sum over each column of $\W_2^{(t)}$, 
and the parameter update is biased toward any directions in $\W_2^{(t)}$ 
on which the gradient has nonzero projection. We refer to \( \W_2^{(t)} \bigl((\W_2^{(t)})^T \nabla_{v}\bigr)\) as \textbf{\emph{adaptive line search}}. Together, the  \textbf{\emph{adaptive line search}} and \textbf{\emph{trainable learning rate}}  suggest that OP-LoRA has an improved ability to navigate complex loss landscapes. While the overall trainable learning rate dynamically updates to the given problem, the adaptive line search can rapidly search the subspace spanned by $W_2$. In Section \ref{sec:mnist_lora}, we explore how this improves performance.

\subsection{MNIST Case Study}
\label{sec:mnist_lora}

To illustrate the advantages of OP-LoRA, we start with a small scale study on MNIST,  showing that OP-LoRA converges to  better loss and also is less sensitive to learning rate. We use a two-layer MLP \(f(x)\) with hidden dimension 512.  We train \(f\) on MNIST for 30 epochs to nearly 0 training loss.  This constitutes our base model, which we freeze before continued LoRA tuning on Rotated MNIST to create an adaptation task. We finetune  the frozen \(f\) with LoRA and OP-LoRA adapters of rank \(r=4\), but this time on the new task of Rotated MNIST, where each MNIST sample is randomly rotated before classification.  

We find two main results. First, we find that OP-LoRA achieves  lower training loss than LoRA. In Figure \ref{fig:mnist_trajectories}, we can see that train loss for LoRA reaches 0.59, vs 0.54 for OP-LoRA.  Second, we find that OP-LoRA is much less sensitive to learning rates than LoRA (Figure \ref{fig:mnist_trajectories} a.), with losses staying relatively low even with learning rates two to three orders of magnitude smaller than optimal.  The learning rate sensitivity of standard LoRA corroborates the findings in ~\citep{biderman2024lora} (See Figure S1  of~\citet{biderman2024lora}).  Qualitatively, we plot the optimization trajectory of both LoRA and OP-LoRA in Figure \ref{fig:mnist_trajectories} b.) and c.). Interestingly, for learning rates which are three orders of magnitude lower than optimal, OP-LoRA makes good progress, demonstrating OP-LoRA's ability to accelerate training. We highlight that this is despite  both methods being optimized with Adam, an optimizer with adaptive learning rates and a momentum term to help with slow learning.

We do additional analysis on the gradient properties of the Hessian using power iteration in Supplementary Materials Section \ref{subsec:structural}, where we find that gradient norm in the direction of highest curvature is much higher for OP-LoRA than standard LoRA. This empirically suggests that OP-LoRA may be less sensitive to poor conditioning than LoRA, since it can minimize loss effectively even when high curvature requires a small learning rate. This is consistent with the view that OP-LoRA adaptively reshapes the LoRA loss landscape for better and faster optimization through reparameterization.

%% file: sec/6_Results.tex
  \begin{table}[t]
      \caption{\textbf{Finetuning Stable Diffusion:} CMMD on Naruto\citep{cervenka2022naruto2} and WikiArt\cite{huggan2023wikiart} (lower is better). }
    \centering
    \vspace{0pt} 
    \setlength{\tabcolsep}{2.2pt}
    \begin{tabular}{lccc}
      \toprule
      \textbf{Method} & \textbf{Naruto } & \textbf{WArt } & \textbf{Avg ($\downarrow$)} \\
      \midrule
      \multicolumn{4}{l}{\textit{W/ grad. alignment}} \\
      LoRA-GA$_{r=4}$\tiny{~\citep{wanglora}}         & \textbf{15.2}  & 43.7 & 29.5  \\
      LoRA-Pro$_{r=4}$\tiny{~\citep{anonymous2025lorapro}}        & 20.9  & 32.4 & 26.7 \\
      SAdamW$_{r=4}$\tiny{~\citep{zhang2024riemannian}}          & 21.9  & \textbf{30.5} &  \textbf{26.2}  \\
      \midrule
      \multicolumn{4}{l}{\textit{W/out grad. alignment}} \\
      PiSSA$_{r=4}$\tiny{~\citep{meng2024pissa}}           & 29.7  & 43.2 & 36.5  \\
      LoRA$_{r=4}$\tiny{~\citep{hu2021lora} }           & 23.7  & 47.9 & 35.8  \\
      DoRA$_{r=4}$\tiny{~\citep{liu2024dora}}          & 17.2  & 49.8 & 33.5  \\
      OP-DoRA$_{r=4}$ (ours)& 11.9 & 46.6 & 29.3  \\
      OP-LoRA$_{r=4}$ (ours)& \textbf{\textbf{9.6}} & \textbf{31.7} & \textbf{\textbf{20.7}}  \\
      \bottomrule
    \end{tabular}
    \label{tab:cmmd_results}
  \end{table}%

\section{Experiments}
\label{sec:experiments}
In this section, we show performance on three different tasks. We start by finetuning Stable Diffusion~\citep{podell2023sdxl} in Section \ref{sec:image_generative}. Low-rank finetuning offers particular advantages in this context, especially since individual users often have only a few images for model specialization, making regularization with low-rank updates particularly valuable. Additionally, storage concerns are significant, as users may want to save numerous specialized model variants. This means that it is beneficial for any over-parameterization scheme to be training only, to minimize storage costs for the end user. In Section \ref{sec:vl}, we move onto visual question answering with a VL-BART~\citep{cho2021unifying} model, where we demonstrate that OP-LoRA and OP-DoRA show consistent improvements over their standard counterparts. Finally, we show results of a finetuned LLaMA~\citep{touvron2023llama} model on Commonsense reasoning tasks in Section \ref{sec:commonsense}. This is another common PEFT use-case, where full finetuning is too expensive and fully-fine tuned models become cumbersome to store.  Finally, in  Supplementary Materials Section \ref{supp:additional_results}, we  show results for improving VeRA~\citep{kopiczkovera}, Mix-of-Show~\citep{gu2023mix}, DreamBooth~\citep{ruiz2023dreambooth} subject-driven generation, and Matrix Factorization with MLP over-parameterization. Details are in the Supplementary Materials.

\vspace{-3pt}
\subsection{Finetuning Stable Diffusion}
\label{sec:image_generative}

We finetune Stable Diffusion XL~\citep{podell2023sdxl} for two  datasets, a Claude Monet subset of Wiki-Art~\citep{huggan2023wikiart} and Blip-Captioned Naruto Images~\citep{cervenka2022naruto2}, and evaluate using MMD distances over CLIP embeddings.\\
\noindent\textbf{Datasets:} \textbf{WikiArt~\citep{huggan2023wikiart}} is a dataset of around 80000 pieces of artwork labeled with artist, genre, and style. We filter by Claude Monet, leaving 1334 images, with text captions of the form `A painting of genre $\langle$$\rangle$'. Importantly, the artist name is not mentioned in the caption, so the style must be learned. \textbf{Naruto BLIP Captions~\citep{cervenka2022naruto2}} is a dataset of 1221 anime images from the Japanese manga series Naruto, captioned by BLIP~\citep{li2022blip}. \\
\noindent\textbf{Finetuning protocol:} We train for two epochs and we  target only the attention layers in the U-Net in Stable Diffusion XL 1.0. We use rank $r=4$ and OP-LoRA MLP width 32. We choose a number of baselines in addition to LoRA intended to make optimization easier. LoRA-GA~\citep{wanglora}, LoRA-Pro~\citep{anonymous2025lorapro} and  ScaledAdamW~\citep{zhang2024riemannian} leverage full finetuning gradient information, while PiSSA~\citep{meng2024pissa} initializes to principal components. DoRA adds a weight normalization to LoRA in order to improve the learning ability of LoRA. We extend OP-LoRA to OP-DoRA by adding an additional prediction head to generate the weight scaling factors alongside the low-rank matrices. 
  \begin{table}[t]
     \caption{\textbf{VQA evaluation with VL-BART:} Measuring accuracy on VQA\cite{goyal2017making}, GQA\cite{hudson2019gqa}, and NLVR\cite{suhr2018corpus}. OP-LoRA and OP-DoRA outperform their non-overparameterized counterparts by around 1\%.}
    \vspace{0pt} 
    \centering
    \setlength{\tabcolsep}{1pt}
    \begin{tabular}{lcccc}
      \toprule
      \textbf{Method} & \textbf{VQAv2} & \textbf{GQA} & \textbf{NLVR} & \textbf{Avg} ($\uparrow$) \\
      \midrule
      Full FT   & 66.9 & 56.7 & 73.7 & 65.8 \\ 
      \midrule
      LoRA$_{r=128}$\tiny{~\citep{hu2021lora}}      & 65.5 & 53.9 & 72.0 & 63.8 \\
      OP-LoRA$_{r=128}$ (ours)  & \textbf{65.6} & \textbf{54.9} & \textbf{73.0} & \textbf{64.5} \\
      \midrule
      DoRA$_{r=128}$\tiny{~\citep{liu2024dora}}      & 65.8 & 54.9 & 72.4 & 64.4 \\
      OP-DoRA$_{r=128}$ (ours)   & \textbf{\textbf{66.4}} & \textbf{\textbf{55.1}} & \textbf{\textbf{74.0}} & \textbf{\textbf{65.2}} \\
      \bottomrule
    \end{tabular}
 
    \label{tab:vl_results}
  \end{table}
\begin{table*}[t!]
\caption{\textbf{Finetuning on commonsense reasoning datasets~\cite{hu2023llm}:}  OP-LoRA and OP-DoRA outperform their standard counterparts on LLaMA-7B~\citep{touvron2023llama}, and are on par with the complex custom optimizers such as LoRA-Pro~\citep{anonymous2025lorapro} and ScaledAdamW~\citep{zhang2024riemannian} despite not leveraging information about the full FT gradient. When combined with ScaledAdamW, OP-LoRA can match standard ScaledAdamW performance with half the inference parameters.}
\small
\centering
\setlength{\tabcolsep}{3pt}
\begin{tabular}{lccccccccccc}
\toprule
\textbf{Method}
  & \makecell{\textbf{Train}\\\textbf{Par.}}
  & \makecell{\textbf{Inf}\\\textbf{Par.}}
  & \textbf{BoolQ}
  & \textbf{PIQA}
  & \textbf{SIQA}
  & \textbf{HSwag}
  & \textbf{WinoG}
  & \textbf{ARC-E}
  & \textbf{ARC-C}
  & \textbf{OBQA}
  & \textbf{AVG} \\
\midrule
\multicolumn{12}{l}{\textit{W/ grad alignment }}\\
LoRA-Pro$_{r=32}$\tiny{~\citep{anonymous2025lorapro}} & 0.83 & 0.83  & 69.6 & 81.6 & 77.7 & \textbf{84.4} & \textbf{80.3} & 81.7 & 65.5 & \textbf{80.2} & 77.6  \\
SAdamW$_{r=32}$\tiny{~\citep{zhang2024riemannian}} & 0.83 & 0.83  & \textbf{70.7} & \textbf{82.3} & \textbf{78.2}  & 83.3 & 79.6 &  \textbf{82.6} & \textbf{66.4}  & 78.6  &  \textbf{77.7}\\
\hdashline
\addlinespace[2pt]  
SAdamW$_{r=16}$\tiny{~\citep{zhang2024riemannian}} & 0.41 & 0.41 & \textbf{69.9} & 81.5 & 77.3 & 82.5 & 79.2 & \textbf{81.4} & 65.2 & 76.8 & 76.7  \\
OP-SAdamW$_{r=16}$(Ours) & 13.5 & 0.41  & \textbf{69.9} & \textbf{81.9} & \textbf{77.8} & \textbf{84.6} & \textbf{81.1} & 80.1 & \textbf{65.5} & \textbf{80.4} & \textbf{77.7} \\

\midrule
\multicolumn{12}{l}{\textit{W/Out grad alignment}}\\
DeepLoRA$_{r=32}$\tiny{~\citep{yaras2024compressible}} & 0.83  & 0.83    & 66.7  & 78.8 & 59.8 & 39.6 & 51.7 &  41.6 & 32.8 & 39.8 & 51.4 \\
HLoader$_{r=32}$\tiny{~\citep{ortiz2024hyperloader}} & 0.83  & 0.83    & 61.5 & 48.5  & 43.5 & 36.3 & 54.3 & 26.2  &  32.1 & 31.8 & 41.8 \\
AdaLoRA$_{r=32}$\tiny{~\citep{zhang2023adalora}} & 1.25 & 0.83  &  67.4 & 80.7 & 77.0 & 47.3 & 79.6 & 81.4&  64.8 & 76.2 & 71.8 \\
DoRA$_{r=32}$\tiny{~\citep{liu2024dora}}  & 0.84  & 0.84  & 65.3 & 65.6 & 76.9 & 81.2 & 78.8 & 79.4 & 64.0 & 78.3 & 73.7 \\
LoRA$_{r=32}$\tiny{~\citep{hu2021lora}} & 0.83 &  0.83  & 67.5 & 80.8 & \textbf{78.2} & 83.4 & 80.4 & 78.0 & 62.6 & 79.1 & 76.3 \\
OP-LoRA(Ours)$_{r=32}$  & 27.4 &  0.83  & \textbf{69.0} & 81.4 & 77.9 & 85.7 & 79.2 & 80.5 & 64.4 & 78.6 & 77.1 \\
OP-DoRA(Ours)$_{r=32}$ & 27.7 & 0.84   & 67.2 & \textbf{82.0} & 76.3 & \textbf{\textbf{86.5}} & \textbf{81.4} & \textbf{81.5} & \textbf{65.3} & \textbf{80.3} & \textbf{77.5} \\
\hdashline
\addlinespace[2pt]  
LoRA$_{r=16}$\tiny{~\citep{hu2021lora}}  & 0.41 &  0.41   & 69.9 &  77.8  & 75.1 & 72.1 & 55.8 &  77.1 & 62.2 & 78.0  & 70.9 \\
OP-LoRA(Ours)$_{r=16}$  & 13.5 & 0.41  & \textbf{70.3}  & \textbf{83.1}  & \textbf{77.7}  & \textbf{77.7}  & \textbf{79.1} & \textbf{78.9} & \textbf{63.9} & \textbf{78.8} & \textbf{76.2} \\
\bottomrule
\end{tabular}
\label{tab:commonsense}
\end{table*}

\noindent\textbf{Evaluation Protocol:} We generate a new image for each training caption and compute the CLIP Maximum Mean Discrepancy~\citep{jayasumana2024rethinking} (CMMD) distance, which measures MMD over CLIP~\cite{radford2021learning} scores. Lower CMMD indicates higher quality generations.~\citet{jayasumana2024rethinking} show CMMD aligns better with human raters than FID or Inception Score. \\
\noindent\textbf{Results:} We present CMMD scores in  
\cref{tab:cmmd_results}. Two interesting trends emerge.  First, OP-LoRA  outperforms OP-DoRA  on both the Naruto and WikiArt datasets.  We attribute this to the increased ease of overfitting with DoRA.  Second,  OP-LoRA and OP-DoRA, achieve  substantially improved scores over their standard counterparts. Specifically, OP-LoRA achieves a CMMD score of 9.6 on the Naruto dataset, compared to 23.7 for LoRA, and similarly, OP-DoRA scores 11.9 compared to DoRA's 17.2. On the WikiArt dataset, OP-LoRA also shows a substantial gain with a score of 31.7 compared to 47.9 for LoRA. Furthermore, OP-LoRA  outperforms other baseline methods on average, including state-of-the-art optimizers such as LoRA-Pro~\citep{anonymous2025lorapro} and ScaledAdamW~\citep{zhang2024riemannian} which leverage information about the full-finetuning gradient. This suggests that following the full-finetuning gradient as closely as possible is not the only way for parameter efficient adapters to perform well, and different approaches are worth exploring. We also show samples of generated images in \cref{fig:qualitative}, where we compare LoRA to OP-LoRA and DoRA to OP-DoRA. We can overall see much higher quality for the over-parameterized variants. For example,  OP-LoRA and OP-DoRA capture the hoodie, while LoRA and DoRA do not. The still life setting for OP-LoRA is more complex, with flowers. Finally, DoRA seems to generate a somewhat degenerate image in the second row, while OP-LoRA and OP-DoRA do not.  We provide many random samples in the Supplementary Materials. We also extend these experiments to DreamBooth~\citep{ruiz2023dreambooth} subject-driven generation in Supplementary Materials Section \ref{sec:dreambooth}, where OP-LoRA improves subject fidelity  over LoRA without sacrificing text alignment.

\vspace{-3pt}
\subsection{Visual Question Answering Experiments}
\label{sec:vl}

\noindent\textbf{Datasets:} \textbf{VQAv2}~\citep{goyal2017making}  and  \textbf{GQA}~\citep{hudson2019gqa}  are both visual question answering datasets.   
\textbf{NLVRv2}~\citep{suhr2018corpus}  is a visual reasoning dataset, answering a True/False question about a pair of images.
\noindent\textbf{Finetuning and evaluation protocol:} We follow~\citet{liu2024dora} in our finetuning protocol. We finetune VL-BART~\citep{pmlr-v139-cho21a} in multi-task way. VL-BART composes a CLIP-ResNET101~\citep{radford2021learning} and Bart~\textsubscript{Base}~\citep{lewis2019bart} model, and trains all tasks jointly with a language-modeling loss. We finetune for 20 epochs with rank $r=128$, targeting only $Q$ and $K$ matrices in attention layers, while also training biases.  
We use MLP hidden layers size 4, based on our experiments in Section \ref{sec:analysis}. 
\noindent\textbf{Results:} We present results in \cref{tab:vl_results}. We can reach a similar conclusion for the vision-language task as for the image generation task; OP-LoRA and OP-DoRA improve over their counterparts. The roughly 1\% improvement achieved by the OP variants matches the gains  from LoRA to DoRA.

\vspace{-3pt}
\subsection{Commonsense Reasoning Experiments}
\label{sec:commonsense}

\noindent\textbf{Datasets:} The Commonsense task consists of 8 sub-tasks, with about 170k training sequences total. \textbf{BoolQ}~\citep{clark2019boolq} is a yes/no question-answering dataset. \textbf{PIQA}~\citep{bisk2020piqa} requires physical knowledge to answer. \textbf{SIQA}~\citep{sap2019socialiqa} is about social reasoning for humans. \textbf{HellaSwag}~\citep{zellers2019hellaswag} asks the model to complete the context with a sentence. \textbf{WinoGrande}~\citep{sakaguchi2021winogrande} asks the model to fill in the blank.  \textbf{ARC-E} and  \textbf{ARC-C}~\citep{clark2018think} are easy and hard variants of multiple choice science questions. \textbf{OBQA}~\citep{mihaylov2018can} asks multiple choice questions requiring strong comprehension skills of context.

 \noindent\textbf{Finetuning and evaluation protocol:} We follow~\citet{liu2024dora} and train with all datasets jointly for 3 epochs, but evaluate each dataset independently. We use $r=32$ and MLP width 32. In addition to baselines intended to make optimization easier by leveraging full finetuning gradient information (LoRA-Pro~\citep{anonymous2025lorapro},  SAdamW~\citep{zhang2024riemannian} and OP-SAdamW), we also compare to the AdaLoRA~\citep{zhang2023adalora}, DeepLoRA~\citep{yaras2024compressible}, and HyperLoader~\citep{ortiz2024hyperloader}. OP-SAdamW adds MLP overparameterization to OP-LoRA by inserting gradient projections from ScaledAdamW to the backward pass into the generating MLP. AdaLoRA dynamically allocates parameters to different layers, useful for PEFT training of large scale models. Like OP-LoRA, DeepLoRA~\citep{yaras2024compressible} is  motivated by overparameterization. In its conception, it trains an over-parameterized adapter and compresses it, but this process is too expensive for large models. Therefore, ~\citet{yaras2024compressible}  simply add a third square matrix to the LoRA adapter, therefore training a product of three matrices. HyperLoader is similar to OP-LoRA, but shares parameters between LoRA-generating MLPs, and is used to test the necessity of decoupling parameters. 
 \begin{figure}[t]
     \captionof{table}{\textbf{GPU Memory and wall time cost}: Evaluated on CommonSense Reasoning with an H100 HBM3 GPU. The increased memory usage is manageable, and Wall Time is faster than alternatives.}
    \centering
    \begin{tabular}{lcc}
      \toprule
      Method     & GPU Mem &  Time   \\
      \midrule
      LoRA\tiny{~\citep{hu2021lora}}       & 44 GB   & 3.5 H  \\
      OP-LoRA                      & 69 GB   & 4 H    \\
      ScaledAdamW\tiny{~\citep{zhang2024riemannian}} & 44 GB   & 4.5 H  \\
      LoRA-Pro\tiny{~\citep{anonymous2025lorapro}}   & 46 GB   & 56 H   \\
      \bottomrule
    \end{tabular}

    \label{tab:method_time}
  \end{figure}%
\textbf{Results} \cref{tab:commonsense} presents the results of our experiments, where both OP-LoRA and OP-DoRA consistently outperform their non-overparameterized counterparts by 1-4\%.  Moreover, OP-LoRA nearly matches LoRA-Pro and ScaledAdamW, but with lower training time (4h vs 56h for LoRA-Pro). Although ScaledAdamW and LoRA-Pro slightly outperform OP-LoRA on commonsense reasoning, they substantially underperform on image generation (Section \ref{sec:image_generative}), and combining ScaledAdamW with OP-LoRA reaches the same performance with \textbf{half the inference parameters due to lower rank}. This enables substantially lower inference storage and serving costs~\citep{sheng2023s}.  Interestingly, HyperLoader, which shares MLP parameters between layers, performs very poorly, supporting our design choice to decouple parameters between LoRA adapters.
\subsection{OP-LoRA Analysis}
\label{sec:analysis}
  \begin{figure}[t]
    \centering
    \includegraphics[width=\linewidth]{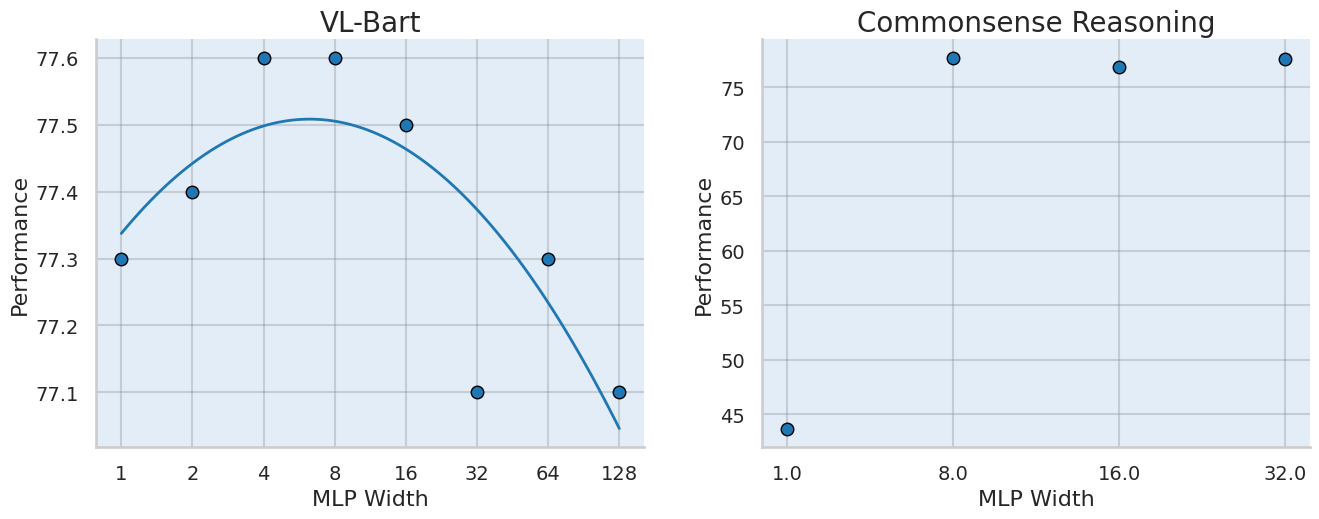}
    \caption{\textbf{Effect of MLP hidden layer size for OP-DoRA:} Performance follows an inverted U-shape for VL-Bart. For Commonsense reasoning, size 1 is too little but otherwise the trend is flat.}
    \label{fig:mlp_width}
  \end{figure}
\noindent\textbf{Computational Costs: }OP-LoRA introduces extra train-time parameters that are discarded at inference, so there’s no added deployment cost. On an H100 HBM3 GPU finetuning LLaMA-7B for 3 epochs with rank = 32, standard LoRA uses 44 GB whereas OP-LoRA uses 69 GB, slowing training by only about 15\%. By contrast,  ScaledAdamW takes around 4.5 h, and LoRA-Pro’s heavier computations extend training to around 56 h. OP-LoRA is around 10\% faster than ScaledAdamW and more than 10x faster than LoRA-Pro. Inference costs can also be decreased; one can achieve the same predictive performance with half the parameters (Table \ref{tab:commonsense}).

\noindent\textbf{Ablating MLP Width:} We study how much to over-parameterize by varying MLP hidden layer size for OP-DoRA on VL-BART and Commonsense Reasoning. In \cref{fig:mlp_width} we see an inverted U-shape for VL-BART; too little width is insufficient but too much degrades performance. For commonsense benchmarks, size 1 is too small but otherwise the trend is flat.

%% file: sec/8_conclusion.tex
\section{Conclusion}
We introduce OP-LoRA, an MLP-based reparameterization of LoRA. By leveraging over-parameterization we accelerate  training without additional inference overhead. Our experiments  across diverse tasks  demonstrate that OP-LoRA consistently improves performance over LoRA, with especially notable gains in image generation where OP-LoRA improves CMMD by up to 15 points. Crucially, the additional parameters are discarded after training, so OP-LoRA incurs zero extra cost at inference time and can even reduce inference costs by matching standard LoRA performance at half the rank. We believe that train-time over-parameterization represents a promising paradigm in model training, and we hope that our work will inspire broader investigation into its applications.

%% file: sec/X_suppl.tex
\setcounter{page}{1}
\maketitlesupplementary

\section{Theoretical Analysis}

\subsection{Extension of overparmeterization analysis to MLP}

Although our analysis of the optimization benefits of OP-LoRA is carried out in the linear case (Section~3.2 of the main paper), the same principles extend to deeper ReLU networks. We formalize this below.

\begin{lemma}[First-order expansion of ReLU activations]
Let $h = \mathrm{ReLU}(a)$ with $a = W_1 z$. 
For a perturbation $\Delta a$, define the activation mask $\mathbf{M} = \mathbf{1}_{\{W_1 z > 0\}}$.  
Then a first-order Taylor expansion gives
\begin{equation*}
h_{\text{new}} \;\approx\; h + \mathbf{M} \odot \Delta a.
\end{equation*}
\end{lemma}

\begin{lemma}[Expansion of pre-activation perturbations]
Let $a = W_1 z$. For small perturbations $\Delta W_1, \Delta z$,
\begin{equation*}
a + \Delta a \;=\; (W_1 + \Delta W_1)(z + \Delta z).
\end{equation*}
Expanding and subtracting $a$ yields
\begin{equation*}
\Delta a = \Delta W_1 z \;+\; W_1 \Delta z \;+\; \Delta W_1 \Delta z.
\end{equation*}
\end{lemma}

\begin{lemma}[Form of gradient updates]
With gradients $\nabla_h, \nabla_z$, the updates take the form
\begin{equation*}
\Delta W_1 = -\eta(\nabla_h \odot \mathbf{M}) z^\top, 
\qquad
\Delta z = -\eta\nabla_z.
\end{equation*}
Substituting into the expansion of $\Delta a$ and dropping second order terms gives
\begin{equation*}
\Delta a
\approx -\eta ((\nabla_h \odot \mathbf{M})\,z^\top z  \;+\; W_1\,\nabla_z).
\end{equation*}
\end{lemma}

\begin{theorem}[Approximation of activation perturbations]
Combining the previous results,
\begin{equation*}
\Delta h \;\approx\; 
-\eta \mathbf{M} \odot \left[(\nabla_h \odot \mathbf{M})\,z^\top z  \;+\; W_1\,\nabla_z\right].
\end{equation*}
\end{theorem}

\begin{theorem}[Update rule in the ReLU case]
Let $v$ be the parameter vector defined as in the main paper.  
Then the update rule is
\begin{align*}
v^{(t+1)} 
&= v^{(t)} 
- \eta\|h^{(t)}\|^2 \nabla_{v^{(t)}} \\
&\quad
- \eta W_2^{(t)}\!\Bigl(
  \mathbf{M} \odot \bigl[(W_2^\top \nabla_v \odot \mathbf{M})\,z^\top z 
  + W_1 \nabla_z \bigr]
\Bigr).
\end{align*}

\end{theorem}

\begin{corollary}[Comparison to the linear case]
In the linear case (Section~3.2 of the main paper), the update reduces to
\begin{equation*}
v^{(t+1)} 
= v^{(t)} 
- \eta\|h^{(t)}\|^2 \nabla_{v^{(t)}} 
- \eta W_2^{(t)} (W_2^{(t)\top} \nabla_v).
\end{equation*}
Thus both cases share two key terms:
\begin{itemize}
    \item \textbf{Trainable learning rate:} $-\|h^{(t)}\|^2 \nabla_v$, unchanged from the linear case.
    \item \textbf{Adaptive line search:} an update along the subspace spanned by the current columns of $W_2$.
\end{itemize}
\end{corollary}

\begin{remark}[Geometric interpretation in the ReLU case]
The adaptive line search retains the same geometric role as in the linear case: shifting updates along the span of $W_2$.  
However, the ReLU nonlinearity induces a \emph{composite over-parameterization}:  
the values $h$ themselves are generated through an extra layer with nonlinearity, and each column of $W_2$ only contributes when its corresponding ReLU unit is active.  
This leads to more diverse update directions when different units activate across inputs or training steps.
\end{remark}
\subsubsection{LoRA Hessian}

We begin with the LoRA reparameterization
\begin{equation*}
\label{eq:lora-def}
W = W_0 + BA,
\qquad
B \in \mathbb{R}^{d \times r},\;
A \in \mathbb{R}^{r \times d},\;
W_0 \in \mathbb{R}^{d \times d}.
\end{equation*}
Let $L(W)$ denote the loss, and let $H_W$ be the Hessian of $L$ at $W_0$, viewed as a linear operator mapping perturbations in $W$ to second–order variations of the loss.

\begin{lemma}[Quadratic approximation]
For any small perturbation $\Delta W$, a second–order Taylor expansion gives
\begin{equation*}
\begin{aligned}
L(W_0 + \Delta W)
&\;\approx\;
L(W_0)
+ \langle \nabla_W L,\, \Delta W \rangle
\\
&\quad
+ \tfrac12 \langle \Delta W,\; H_W(\Delta W)\rangle.
\end{aligned}
\end{equation*}
\end{lemma}

\begin{theorem}[Effective Hessian with respect to $B$]
Fix $A$ and consider variations in $B$.  
For a perturbation $\Delta B$, we have $\Delta W = \Delta B A$.  
The corresponding effective curvature operator is
\begin{equation*}
H_B(\Delta B) \;=\; H_W(\Delta B A)\,A^\top.
\end{equation*}
\end{theorem}

\begin{proof}
Substitute $\Delta W = \Delta B A$ into the quadratic form:
\begin{align*}
\tfrac12 \langle \Delta W,\; H_W \Delta W\rangle
&= \tfrac12 \langle \Delta B A,\, H_W(\Delta B A)\rangle \\
&= \tfrac12 \langle \Delta B,\, H_W(\Delta B A)\,A^\top \rangle,
\end{align*}
which establishes the claim.
\end{proof}

\begin{corollary}[Operator and matrix forms of $H_B$]
In operator notation,
\[
H_B \;=\; (\cdot\,A^\top)\,\circ\,H_W\,\circ\,(\cdot A),
\]
where $(\cdot X)$ denotes right multiplication by $X$.  
In matrix form,
\begin{equation*}
H_B \;=\; (A^\top \otimes I)^\top\,H_W\,(A^\top \otimes I), 
\qquad I \in \mathbb{R}^{d \times d}.
\end{equation*}
\end{corollary}

\begin{theorem}[Effective Hessian with respect to $A$]
Fix $B$ and consider variations in $A$.  
For a perturbation $\Delta A$, we have $\Delta W = B \Delta A$.  
The corresponding effective curvature operator is
\begin{equation*}
H_A(\Delta A) \;=\; B^\top H_W(B \Delta A).
\end{equation*}
\end{theorem}

\begin{proof}
Substitute $\Delta W = B \Delta A$ into the quadratic form:
\begin{align*}
\tfrac12 \langle \Delta W,\; H_W \Delta W\rangle
&= \tfrac12 \langle B \Delta A,\, H_W(B \Delta A)\rangle \\
&= \tfrac12 \langle \Delta A,\, B^\top H_W(B \Delta A)\rangle,
\end{align*}
which establishes the claim.
\end{proof}

\begin{corollary}[Operator and matrix forms of $H_A$]
In operator notation,
\[
H_A \;=\; B^\top \,\circ\, H_W \,\circ\, B.
\]
In matrix form,
\begin{equation*}
H_A \;=\; (I \otimes B^\top)\,H_W\,(I \otimes B),
\qquad I \in \mathbb{R}^{d \times d}.
\end{equation*}
\end{corollary}

\subsubsection{Condition number bounds}

In the main paper we state the following bound on the condition number of $H_A$ :
\begin{equation*}
\label{eq:HA-main-bound}
\frac{\kappa(B)^2}{\kappa(H_W)}
\;\leq\;
\kappa(H_A)
\;\leq\;
\kappa(H_W)\,\kappa(B)^2,
\end{equation*}
where $\kappa(\cdot)$ denotes the spectral condition number. We first prove a general bound for quadratic forms of the type
$S^\top H_W S$, then specialize to the LoRA case $S = I \otimes B$.

\begin{lemma}[Spectral bounds for $S^\top H_W S$]
\label{lem:quadratic-bounds}
Let $H_W \in \mathbb{R}^{n\times n}$ be symmetric positive definite (SPD),
and let $S \in \mathbb{R}^{n\times m}$ be full column rank.
Denote by $\lambda_{\min}(\cdot)$ and $\lambda_{\max}(\cdot)$
the minimal and maximal eigenvalues, and by
$\sigma_{\min}(\cdot)$ and $\sigma_{\max}(\cdot)$ the minimal and maximal singular values.
Define
\[
H_S \;=\; S^\top H_W S.
\]
Then
\begin{align*}
\lambda_{\min}(H_W)\,\sigma_{\min}(S)^2
&\;\leq\;
\lambda_{\min}(H_S)
\;\leq\;
\lambda_{\max}(H_W)\,\sigma_{\min}(S)^2, 
\\[0.3em]
\lambda_{\min}(H_W)\,\sigma_{\max}(S)^2
&\;\leq\;
\lambda_{\max}(H_S)
\;\leq\;
\lambda_{\max}(H_W)\,\sigma_{\max}(S)^2. 
\end{align*}
\end{lemma}

\begin{proof}
For any nonzero $x \in \mathbb{R}^m$,
\[
x^\top H_S x
= x^\top S^\top H_W S x
= (Sx)^\top H_W (Sx).
\]
Since $H_W$ is SPD,
\[
\lambda_{\min}(H_W)\,\|Sx\|^2
\;\le\;
(Sx)^\top H_W (Sx)
\;\le\;
\lambda_{\max}(H_W)\,\|Sx\|^2.
\]
Using the singular value bounds
$\sigma_{\min}(S)\,\|x\| \le \|Sx\| \le \sigma_{\max}(S)\,\|x\|$,
we obtain, for all nonzero $x$,
\[
\begin{aligned}
\lambda_{\min}(H_W)\,\sigma_{\min}(S)^2\,\|x\|^2
&\;\le\;
x^\top H_S x
\\
&\;\le\;
\lambda_{\max}(H_W)\,\sigma_{\max}(S)^2\,\|x\|^2.
\end{aligned}
\]

Taking the minimum over unit vectors $x$ gives
\[
\begin{aligned}
\lambda_{\min}(H_W)\,\sigma_{\min}(S)^2
&\;\leq\;
\lambda_{\min}(H_S)
\\
&\;\leq\;
\lambda_{\max}(H_W)\,\sigma_{\min}(S)^2,
\end{aligned}
\]
since $\|Sx\| \ge \sigma_{\min}(S)\|x\|$ holds for all $x$.

Taking the maximum over unit vectors $x$ gives
\[
\begin{aligned}
\lambda_{\min}(H_W)\,\sigma_{\max}(S)^2
&\;\leq\;
\lambda_{\max}(H_S)
\\
&\;\leq\;
\lambda_{\max}(H_W)\,\sigma_{\max}(S)^2,
\end{aligned}
\]
since $\|Sx\| \le \sigma_{\max}(S)\|x\|$ holds for all $x$.
This establishes the bounds.

\end{proof}

\begin{corollary}[Condition number of $S^\top H_W S$]
\label{cor:quadratic-kappa}
Under the assumptions of Lemma~\ref{lem:quadratic-bounds},
\[
\frac{\kappa(S)^2}{\kappa(H_W)}
\;\leq\;
\kappa(H_S)
\;\leq\;
\kappa(H_W)\,\kappa(S)^2,
\]
where $\kappa(H_S) = \lambda_{\max}(H_S)/\lambda_{\min}(H_S)$
and $\kappa(S) = \sigma_{\max}(S)/\sigma_{\min}(S)$.
\end{corollary}

\begin{proof}
From Lemma~\ref{lem:quadratic-bounds}, we have
\begin{align*}
\lambda_{\max}(H_S)
&\;\leq\;
\lambda_{\max}(H_W)\,\sigma_{\max}(S)^2,\\
\lambda_{\min}(H_S)
&\;\geq\;
\lambda_{\min}(H_W)\,\sigma_{\min}(S)^2.
\end{align*}
Therefore
\begin{align*}
\kappa(H_S)
&= \frac{\lambda_{\max}(H_S)}{\lambda_{\min}(H_S)} \\
&\;\leq\;
\frac{\lambda_{\max}(H_W)\,\sigma_{\max}(S)^2}
     {\lambda_{\min}(H_W)\,\sigma_{\min}(S)^2} \\
&= \kappa(H_W)\,\kappa(S)^2.
\end{align*}

Similarly, Lemma~\ref{lem:quadratic-bounds} also gives
\begin{align*}
\lambda_{\max}(H_S)
&\;\geq\;
\lambda_{\min}(H_W)\,\sigma_{\max}(S)^2,\\
\lambda_{\min}(H_S)
&\;\leq\;
\lambda_{\max}(H_W)\,\sigma_{\min}(S)^2,
\end{align*}
so
\begin{align*}
\kappa(H_S)
&= \frac{\lambda_{\max}(H_S)}{\lambda_{\min}(H_S)} \\
&\;\geq\;
\frac{\lambda_{\min}(H_W)\,\sigma_{\max}(S)^2}
     {\lambda_{\max}(H_W)\,\sigma_{\min}(S)^2} \\
&= \frac{\kappa(S)^2}{\kappa(H_W)}.
\end{align*}
This establishes the claimed bounds.
\end{proof}

We now specialize to the LoRA Hessians $H_A$ and $H_B$.

\begin{theorem}[Condition number bounds for $H_A$]
\label{thm:HA-bounds}
Assume $H_W$ is SPD and $B$ is full rank.
Let $H_A$ be the effective Hessian with respect to $A$,
\[
H_A = (I \otimes B^\top)\,H_W\,(I \otimes B).
\]
Then
\begin{equation*}
\frac{\kappa(B)^2}{\kappa(H_W)}
\;\leq\;
\kappa(H_A)
\;\leq\;
\kappa(H_W)\,\kappa(B)^2.
\end{equation*}
\end{theorem}

\begin{proof}
Set $S = I \otimes B$, so that $H_A = S^\top H_W S$.
The singular values of $I \otimes B$ are exactly the singular values of $B$,
repeated, so $\kappa(S) = \kappa(B)$.
Applying Corollary~\ref{cor:quadratic-kappa} with this choice of $S$
yields the stated bounds.
\end{proof}

\begin{corollary}[Condition number bounds for $H_B$]
\label{cor:HB-bounds}
Assume $H_W$ is SPD and $A$ is full rank.
Let $H_B$ be the effective Hessian with respect to $B$,
\[
H_B = (A^\top \otimes I)^\top\,H_W\,(A^\top \otimes I).
\]
Then
\begin{equation*}
\frac{\kappa(A)^2}{\kappa(H_W)}
\;\leq\;
\kappa(H_B)
\;\leq\;
\kappa(H_W)\,\kappa(A)^2.
\end{equation*}
\end{corollary}

\begin{proof}
Apply Corollary~\ref{cor:quadratic-kappa} with $S = A^\top \otimes I$,
using again that $\kappa(A^\top \otimes I) = \kappa(A)$.
\end{proof}

\section{Additional Results}
\label{supp:additional_results}

 \begin{figure*}[t]
    \centering
    \begin{minipage}{0.24\textwidth}
        \centering
        \includegraphics[width=\textwidth]{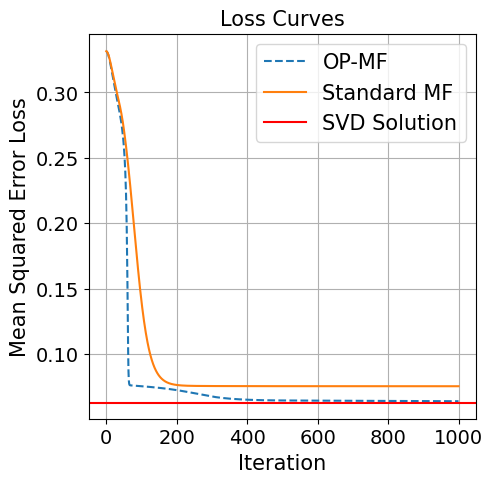}
        \subcaption{\textbf{Loss Curve}} 
    \end{minipage}%
    \hfill
    \begin{minipage}{0.24\textwidth}
        \centering
        \includegraphics[width=0.95\textwidth]{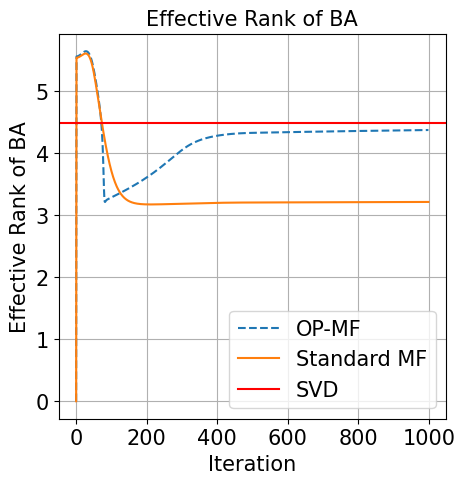}
        \subcaption{\textbf{Effective Rank}} 
    \end{minipage}
    \hfill
    \begin{minipage}{0.24\textwidth}
        \centering
        \includegraphics[width=\textwidth]{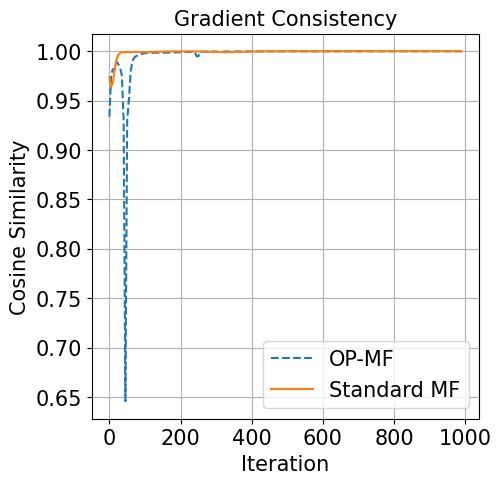}
        \subcaption{\textbf{Gradient Consistency} } 
    \end{minipage}%
    \hfill
    \begin{minipage}{0.24\textwidth}
        \centering
        \includegraphics[width=\textwidth]{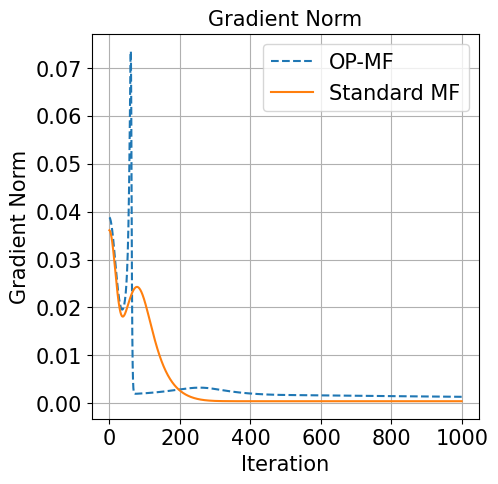}
        \subcaption{\textbf{Gradient Norm}} 
    \end{minipage}
    \caption{\textbf{Matrix Factorization (MF) Gradient Analysis:} \textbf{(a)} Loss Curve shows the reconstruction error for matrix factorization. OP-MF converges better and faster than standard MF. \textbf{(b)} Effective Rank of BA reveals changes in the rank  of the learned solution. OP-MF learns an effective rank closer to that of the ground-truth SVD solution. \textbf{(c)} Gradient Consistency measures the similarity of gradients across iterations. OP-MF is able to make a sudden change in optimization direction, while standard MF cannot. \textbf{(d)} Gradient Norm illustrates the scale of gradients. OP-MF is able more quickly adjust optimization step size. }  
    \label{fig:gradients}
\end{figure*}

\subsection{Gradient Analysis LoRA and OP-LoRA}
\label{subsec:structural}

Recall that in  the main paper we show
\begin{equation*}
\label{eq:cond-bounds}
\frac{\kappa(B)^2}{\kappa(H_W)}
\;\leq\;
\kappa(H_A)
\;\leq\;
\kappa(H_W)\,\kappa(B)^2,
\end{equation*}
where $A$ and $B$ are LoRA matrices, $W$ are the base weights, $\kappa(\cdot)$ denotes the condition number, and $H$ are the corresponding Hessians. Equation ~\eqref{eq:cond-bounds} implies that LoRA can exhibit worse Hessian conditioning than full finetuning.

A high Hessian condition number indicates very high curvature in some directions relative to others. This matters because the step size must be small enough to avoid instabilities along the highest-curvature direction; the maximal stable learning rate is then limited by that direction and may be too small to make progress in low-curvature directions. This becomes problematic when the highest-curvature direction has already been minimized, but the low-curvature directions still require larger learning rates.

A useful diagnostic is the magnitude of the gradient in the direction of the principal singular vector of the Hessian of the trainable parameters. Let $v$ be that direction (estimated by power iteration) and $g$ the gradient. If $|v^\top g|$ is relatively large, the loss can still be reduced even with a learning rate small enough to remain stable in the largest-curvature direction. Conversely, if $|v^\top g|\approx 0$, a large condition number will prevent further decrease of the loss.

\paragraph{Setup.}
We use the power-iteration method to estimate $v$ and then measure $|v^\top g|$ for a small-scale Rotated-MNIST problem. Pre-training is performed on MNIST and continued training is on Rotated MNIST (as in Fig.~2 of the main paper). We report the terminal values for LoRA and OP-LoRA.

\begin{table}[ht]
\centering
\small
\begin{tabular}{lc}
\toprule
Method & $|v^\top g|$ \\
\midrule
OP-LoRA & 0.42 \\
LoRA    & 0.06 \\
\bottomrule
\end{tabular}
\caption{$|v^\top g|$ at the end of training on Rotated MNIST. Higher is better (indicates remaining descent along the highest-curvature direction).}
\label{tab:vtg}
\end{table}

\paragraph{Findings.}
OP-LoRA exhibits a much larger $|v^\top g|$ in the direction of largest curvature than LoRA (Table~\ref{tab:vtg}). Empirically, this suggests OP-LoRA may be \emph{less sensitive} to poor conditioning than LoRA, because it can continue to reduce the loss even when the step size is constrained by the highest-curvature direction. Thus, even if the OP-LoRA MLP were itself poorly conditioned, this sensitivity matters less than for LoRA.

These observations are consistent with the view that OP-LoRA adaptively reshapes the LoRA loss landscape via reparameterization, leading to better and faster optimization.

\subsection{A Matrix Factorization Case Study}
\label{subsec:mf}

To verify that the gradient properties of MLP over-parameterization results in observable changes, we design a controlled matrix factorization experiment comparing MLP-generated low-rank matrices \( A \) and \( B \) with freely learned parameter matrices and measure convergence and gradients. 

Matrix factorization decomposes a target matrix \( M \in \mathbb{R}^{m \times n} \) into two lower-dimensional matrices, \( A \in \mathbb{R}^{r \times n} \) and \( B \in \mathbb{R}^{m \times r} \), where \( r \) is the latent dimension or rank. This decomposition allows us to approximate \( M \) by \( BA \).  It can be solved exactly with SVD, or as in our study, one can use gradient descent to minimize the reconstruction error: 

\[
 \| M - BA \|_F^2,
\]
where \( \| \cdot \|_F \) denotes the Frobenius norm. This resembles LoRA-tuning, where the pre-trained base weights are set to all zeros and the target matrix $M$ is the full  finetuning gradient matrix, making it an interesting proxy problem to study. 

\noindent\textbf{Experimental setup and training protocol:}  We construct a synthetic target matrix \( M \in \mathbb{R}^{100 \times 100} \) with entries initialized uniformly at random from 0 to 1. The resulting matrix has a poor condition number, defined as the ratio between the largest singular value and lowest, making the optimization difficult and therefore a good test for MLP reparameterization.  We train for 1000 steps with SGD, with linear warmup  for 50 steps and linear learning rate decay. 

\noindent\textbf{OP-MF Model:}  The OP-MF model generates matrices \( A \) and \( B \) through two separate MLPs. We enforce both matrices to be of rank 8. Each MLP receives a learned input vector \( z \in \mathbb{R}^{128} \) and processes it through two fully connected layers with 32 hidden units and ReLU activations, outputting the entries for either \( A \) or \( B \). The second layer is heavily overparameterized; the parameter count is number of hidden units in the MLP by the size of the parameter matrix $A$ or $B$. To align with LoRA’s initialization strategy, the MLP for matrix \( B \) is initialized to output zeros, setting the model close to a pre-trained state. \\
\noindent\textbf{Matrix Factorization(MF) Model:}  We train a MF model with freely learnable matrices \( A \) and \( B \), initialized with random values for \( A \) and zeros for \( B \). Again, both matrices have rank 8. 

\noindent\textbf{Finding 1: OP-MF rapidly adapts step size and direction.} We examine the gradients, looking for evidence that OP-LoRA adaptively changes step size and direction. Our results reveal that as predicted,  OP-MF shows an ability to rapidly adapt step sizes. This can be seen in \cref{fig:gradients} (d), where the gradient norm experiences a sharp spike, followed by a collapse, corresponding to acceleration and slow down in the loss curve in \cref{fig:gradients} (a). The sudden phase change in gradient norm also corresponds to a direction change in trajectory, measured by the cosine similarity between gradients at 10-step intervals in in \cref{fig:gradients} (c). Therefore, MLP reparameterization rapidly changes step size and direction, as suggested by the mathematical analysis in Section 3 of the main paper. 

\noindent\textbf{Finding 2: OP-MF is more effective at reaching the SVD solution than standard MF trained with SGD.} In \cref{fig:gradients}(a), we study the loss curves for both the MF model and the OP-MF model. The plot tracks MSE loss over 1000 iterations. The red line represents the SVD solution as a baseline. Interestingly, OP-MF solutions reach the best-case reconstruction error achieved with SVD, while Standard-MF cannot. 

In addition to reconstruction error, another way to track progress towards a solution in matrix factorization is plotting the effective rank of the predicted matrix \( BA \) over the course of training. Effective rank $\rho$ is defined as     
\[
 \rho = \exp\left(- \sum_{i=1}^{r} \sigma_i \log \sigma_i \right)
 \]
where \( \sigma_i \) represents the normalized singular values of \( BA \), and \( r \) is the rank of the matrix. One would expect the effective rank to converge towards the effective rank of the ground-truth SVD solution for successful optimization runs.

In \cref{fig:gradients} (b), we observe the behavior of effective rank across iterations for both MF and OP-MF. We find that OP-MF can approximate the effective rank of the best-case SVD solution much more closely than standard MF. 

\begin{figure*}[ht!]
    \centering
    \begin{minipage}{0.32\textwidth}
        \centering
        \includegraphics[width=\textwidth]{Figs/Loss.png}
        \subcaption{SGD} 
    \end{minipage}%
    \hfill
    \begin{minipage}{0.32\textwidth}
        \centering
        \includegraphics[width=\textwidth]{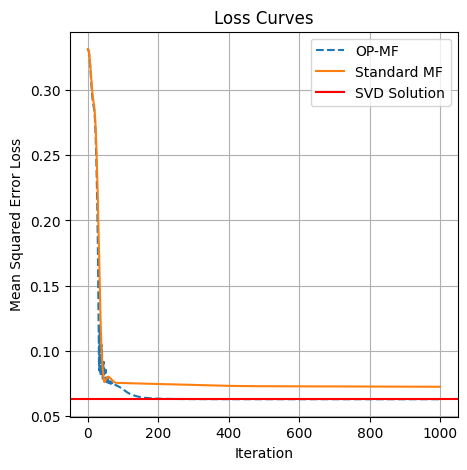}
        \subcaption{Momentum} %
    \end{minipage}%
    \hfill
    \begin{minipage}{0.32\textwidth}
        \centering
        \includegraphics[width=\textwidth]{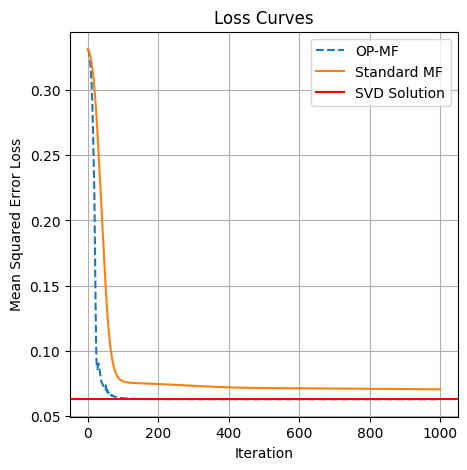}
        \subcaption{Adam} 
    \end{minipage}
    \caption{\textbf{Comparing Optimizers.} MLP reparameterization accelerates matrix factorization, even with advanced optimizers.}
    \label{fig:optimizers}
\end{figure*}

\noindent\textbf{Finding 3: OP-MF composes well with both SGD with Momentum and Adam.} A natural question is if using standard  acceleration methods like SGD with Momentum or Adam is enough. In Figure \ref{fig:optimizers} we present experiments by adding Momentum to SGD and replacing it entirely with Adam, an optimizer that combines adaptive learning rates with momentum .  Both Adam and SGD with Momentum improve reconstruction error for MF, but neither reach the SVD solution. Moreover, OP-MF composes with even best-case and advanced optimizers to find the SVD solution even faster, indicating their complimentary nature.

\subsection{DreamBooth Subject-Driven Generation}
\label{sec:dreambooth}

We extend our image generation experiments to DreamBooth~\citep{ruiz2023dreambooth}, finetuning SDXL and evaluating on the DreamBench benchmark, which consists of 30 diverse subjects, 25 prompts per subject, and 4 generations per prompt. We report CLIP-I (semantic similarity via CLIP embeddings), DINO (fine-grained visual similarity via self-supervised features), and CLIP-T (text-image alignment).

\begin{table}[ht]
\centering
\begin{tabular}{lcc}
\toprule
Metric & OP-LoRA & LoRA \\
\midrule
CLIP-I (subject fidelity) & \textbf{0.83} & 0.82 \\
DINO (subject fidelity) & \textbf{0.69} & 0.65 \\
CLIP-T (text alignment) & 0.24 & \textbf{0.25} \\
\bottomrule
\end{tabular}
\caption{DreamBench results with SDXL. OP-LoRA improves subject fidelity (DINO) without sacrificing text alignment.}
\label{tab:dreambench}
\end{table}

\begin{figure}[ht]
\centering
\begin{tabular}{@{}c@{\hspace{1pt}}c@{\hspace{1pt}}c@{}}
\includegraphics[width=0.3\linewidth]{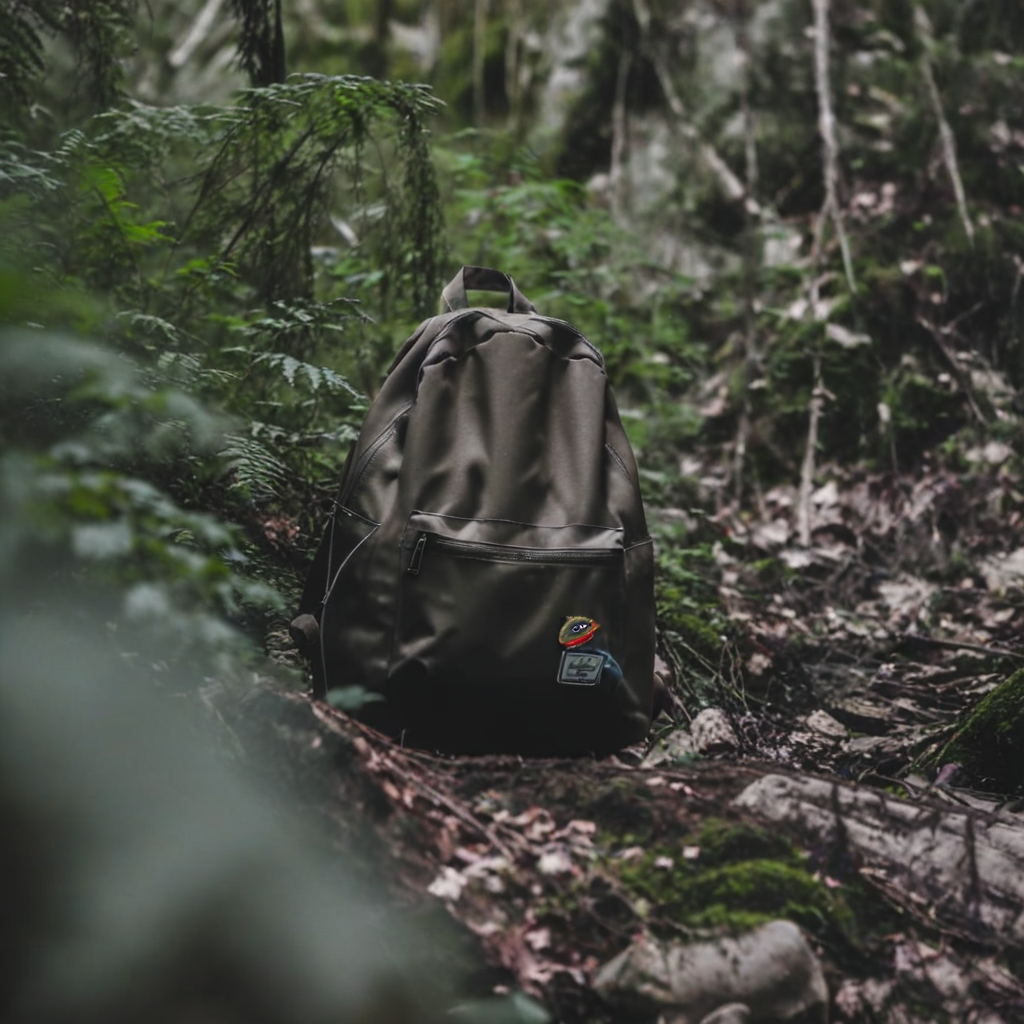} &
\includegraphics[width=0.3\linewidth]{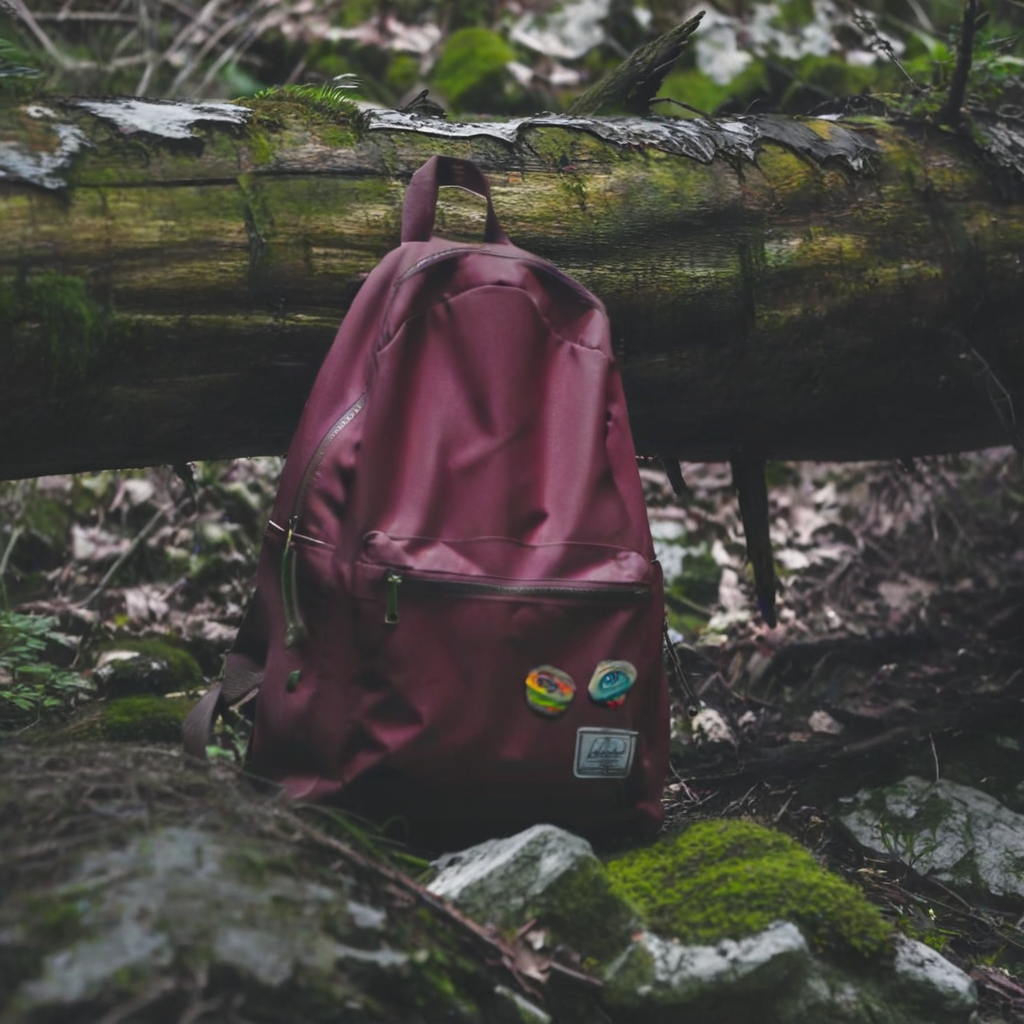} &
\includegraphics[width=0.3\linewidth]{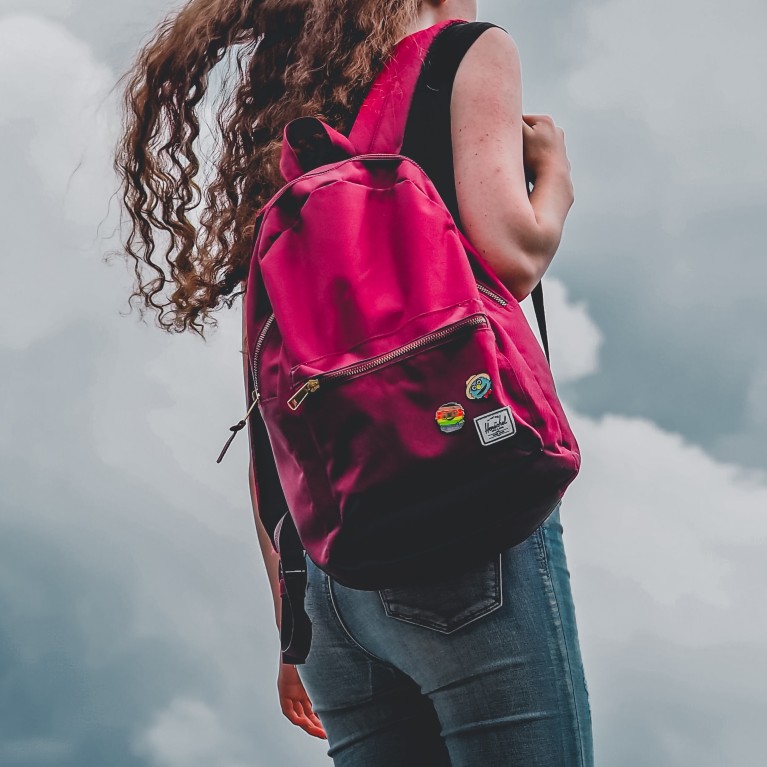} \\[-3pt]
\scriptsize (a) LoRA & \scriptsize (b) OP-LoRA & \scriptsize (c) Reference subject
\end{tabular}
\caption{DreamBench: OP-LoRA preserves subject identity better, with more accurate color and patches. Prompt: \textit{A [V] backpack in jungle.}}
\label{fig:dreambench}
\end{figure}

OP-LoRA outperforms LoRA on fine-grained visual similarity (DINO: 0.69 vs 0.65), without sacrificing performance on more global CLIP semantic similarity and text-image alignment. Qualitatively, OP-LoRA better preserves subject details like color and patches (\cref{fig:dreambench}).

\subsection{LLaVA Image Classification}
\label{sec:classifciation}
\citet{zhang2024visually} recently demonstrated that by applying simple finetuning to adapters, large multimodal models like LLaVA can achieve surprisingly high performance on a variety of classification tasks. We leverage this finding, replacing full finetuning with LoRA and OP-LoRA.\\
\noindent\textbf{Datasets:} We use the Stanford Cars \citep{krause20133d} dataset, which is a fine-grained dataset of about 8000 training examples consisting of 196 classes of cars.

\noindent\textbf{finetuning and evaluation protocol:} We follow \citep{zhang2024visually}, and convert image classification into a captioning task by using the format ``$\langle$ image$\rangle$ What type of object is in this photo? $\langle$ class name $\rangle$'' and training with a language modeling objective. We finetune the visual projector layers of LLaVA1.5-7B for 50 epochs. We set MLP width to 768, since memory resources are not an issue for training only the adapter. At evaluation, we parse the generations and search for the correct class label. 

\noindent\textbf{Results:} We present the results in \cref{tab:llava_performance}. OP-LoRA consistently outperforms LoRA at both rank levels, achieving 83.6\% at rank 16 and 87.1\% at rank 64, compared to 82.8\% and 86.3\% with LoRA. This result further reinforces the efficacy of MLP re-parameterization.

\begin{table}[ht!]
\centering
\begin{tabular}{lcc}
\toprule
\textbf{Rank} & \textbf{LoRA} & \textbf{OP-LoRA} \\
\midrule
16 & 82.8 & \textbf{83.6} \\
64 & 86.3 & \textbf{87.1} \\
\bottomrule
\end{tabular}
\caption{LLaVA1.5-7B Image Classification, Top-1 Accuracy on Stanford Cars~\citep{krause20133d}.}
\label{tab:llava_performance}
\end{table}

\subsection{Stability of OP-DoRA}

We found that OP-DoRA is more stable than DoRA, as shown with standard deviations across 3 runs. We hypothesize that this is a reflection of decreased learning rate sensitivity. 

\begin{table}[ht]
 \centering
 \begin{tabular}{l c}
   \toprule
   Method                 & Commonsense          \\
   \midrule
   DoRA ($r=32$)          & $73.7 \pm 6.7$       \\
   OP-DoRA ($r=32$)       & $77.5 \pm 1.6$       \\
   \bottomrule
 \end{tabular}
\end{table}

\subsection{ VeRA and OP-VeRA}

We extend our method to \textsc{VeRA}~\citep{kopiczkovera}, an ultra–low-parameter variant of LoRA, and evaluate on the GLUE\citep{wang2018glue} benchmark using a RoBERTa-base\citep{liu2019roberta} backbone following the setup in~\citep{kopiczkovera}. We report results on \textsc{SST-2}, \textsc{CoLA}, and \textsc{QNLI}. We exclude \textsc{MNLI} due to computational constraints and therefore also exclude\textsc{MRPC}/\textsc{RTE}/\textsc{STS-B}, which are commonly initialized from \textsc{MNLI} to mitigate overfitting~\citep{hu2021lora}.

\begin{table}[ht]
\centering
\small
\begin{tabular}{lcccc}
\toprule
 & \textsc{SST-2} & \textsc{CoLA} & \textsc{QNLI} & \textbf{Avg.} \\
\midrule
VeRA & 94.0 & 59.8 & \textbf{91.7} & 81.8 \\
OP\mbox{-}VeRA & \textbf{94.3} & \textbf{62.1} & 91.5 & \textbf{82.6} \\
\bottomrule
\end{tabular}
\caption{GLUE dev results with RoBERTa-base.}
\label{tab:op-vera}
\end{table}

Averaged over the three tasks, OP\mbox{-}VeRA improves upon VeRA by \textbf{1.2} points, indicating the generality of the proposed optimization.

\subsection{Improving Mix-of-Show with OP-LoRA}

\begin{figure*}[ht!]
    \centering
    \includegraphics[width=0.8\textwidth]{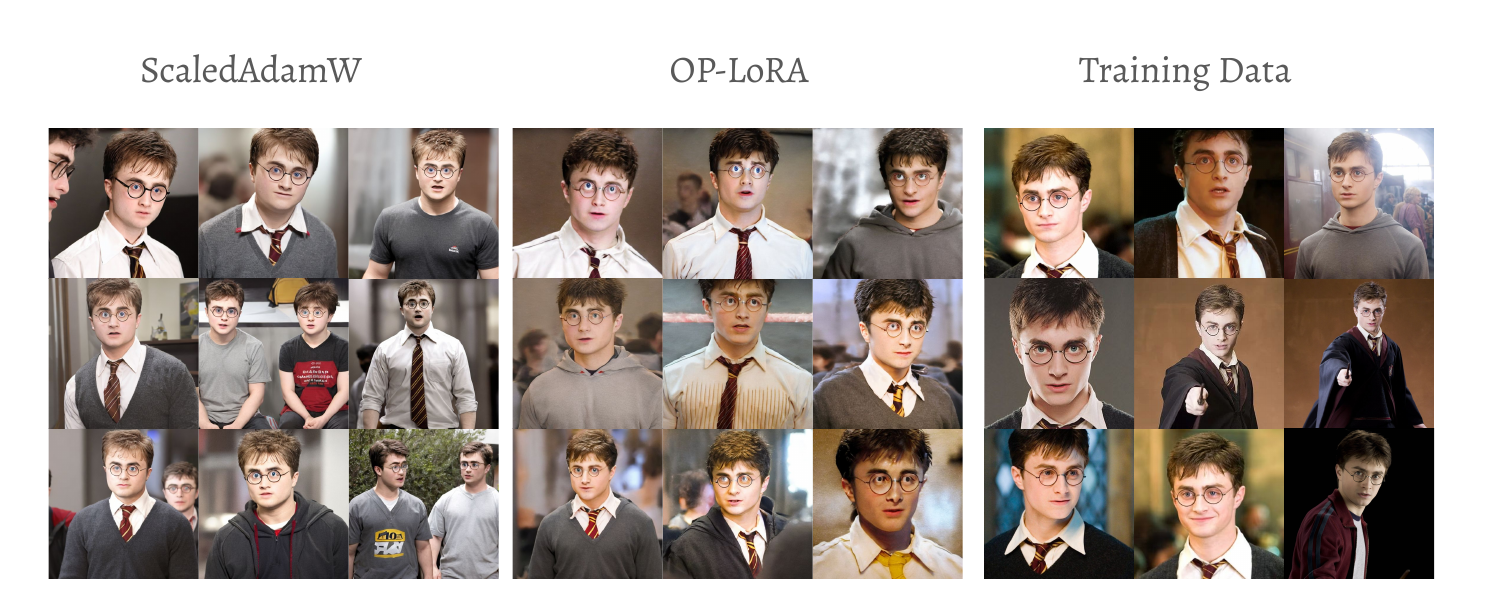}
    \caption{OP-LoRA vs ScaledAdamW\citep{zhang2024riemannian} with Mix-of-Show\citep{gu2023mix}. }
    \label{fig:mixofshow}
\end{figure*}

Following \citet{zhang2024riemannian} directly, we apply Mix-of-Show to a small set of 14 training images of Harry Potter. We compare standard ScaledAdamW and OP-LoRA, keeping all training settings from \citet{zhang2024riemannian}. In Figure \ref{fig:mixofshow}, we generate images from the learned <potter> tokens as a prompt. We see that OP-LoRA better captures the subject of Harry Potter; there are fewer image of two people (not present in the training set) and the shape of the face is more accurate. Moreover, OP-LoRA generates Harry Potter in causal clothing less frequently.  
 
\subsection{Qualitative Stable Diffusion Results }

\begin{figure*}
    \centering
    \includegraphics[width=0.9\textwidth]{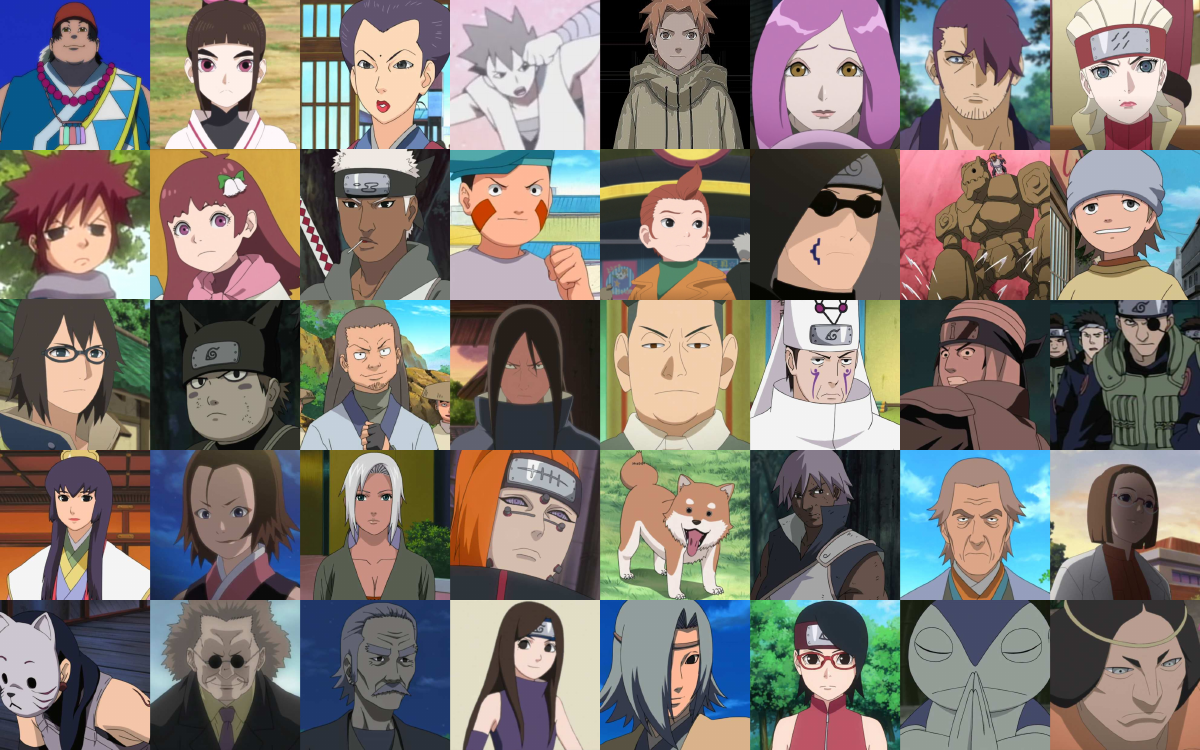}
    \caption{GT Naruto\citep{cervenka2022naruto2} images}
    \label{fig:naruto_gt}
\end{figure*}

In the main paper, we show the large quantitative gains OP-LoRA/OP-DoRA give in the image generation task (Table 1). We now present extensive random generations in Figures \ref{fig:naruto_gt} though \ref{fig:monet_scaledadamw}, with captions from the dataset as input. Several general trends emerge. First, there is a strong color bias towards red, however OP-LoRA and OP-DoRA reduce this dramatically. We attibute this improvement to the over-parameterization easing optimization.  Second, over-parameterized LoRA is generates much more diverse and more complex scenes. Overall, qualitative results match the quantitative metrics.

\begin{figure*}
    \centering
    \includegraphics[width=0.9\textwidth]{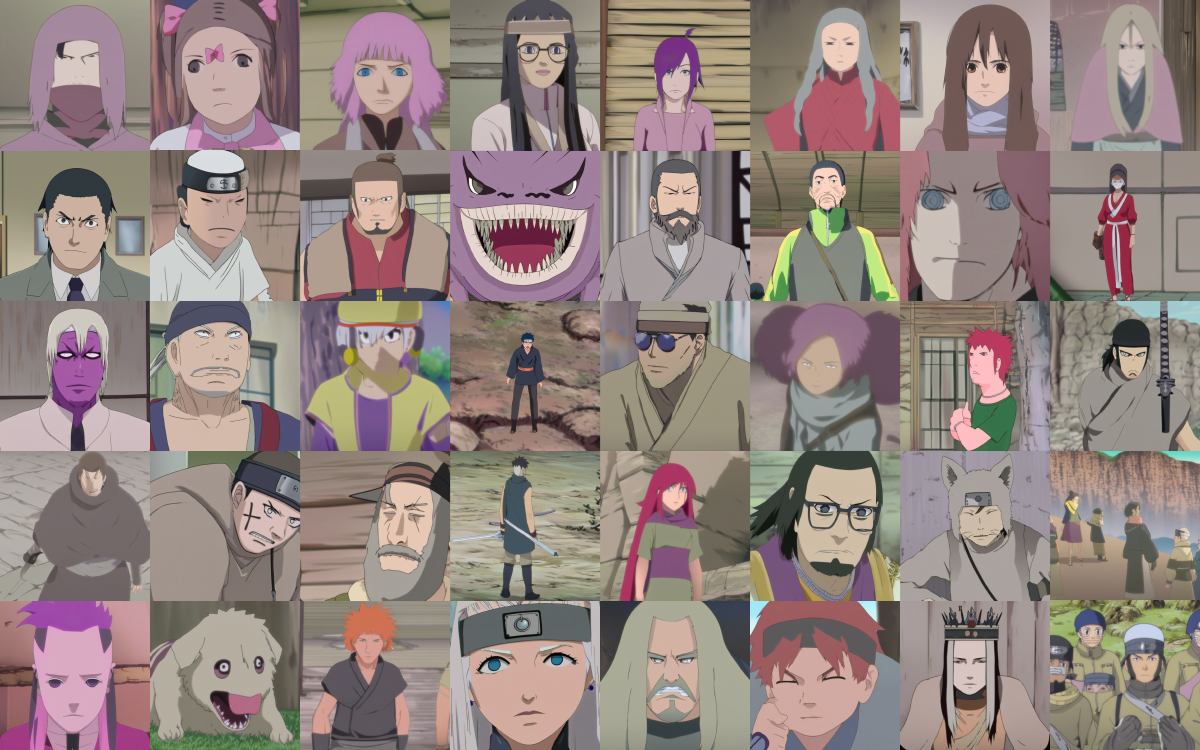}
    \caption{ OP-LoRA Naruto\citep{cervenka2022naruto2} generated images. }
    \label{fig:naruto_oplora}
\end{figure*}

\begin{figure*}
    \centering
    \includegraphics[width=0.9\textwidth]{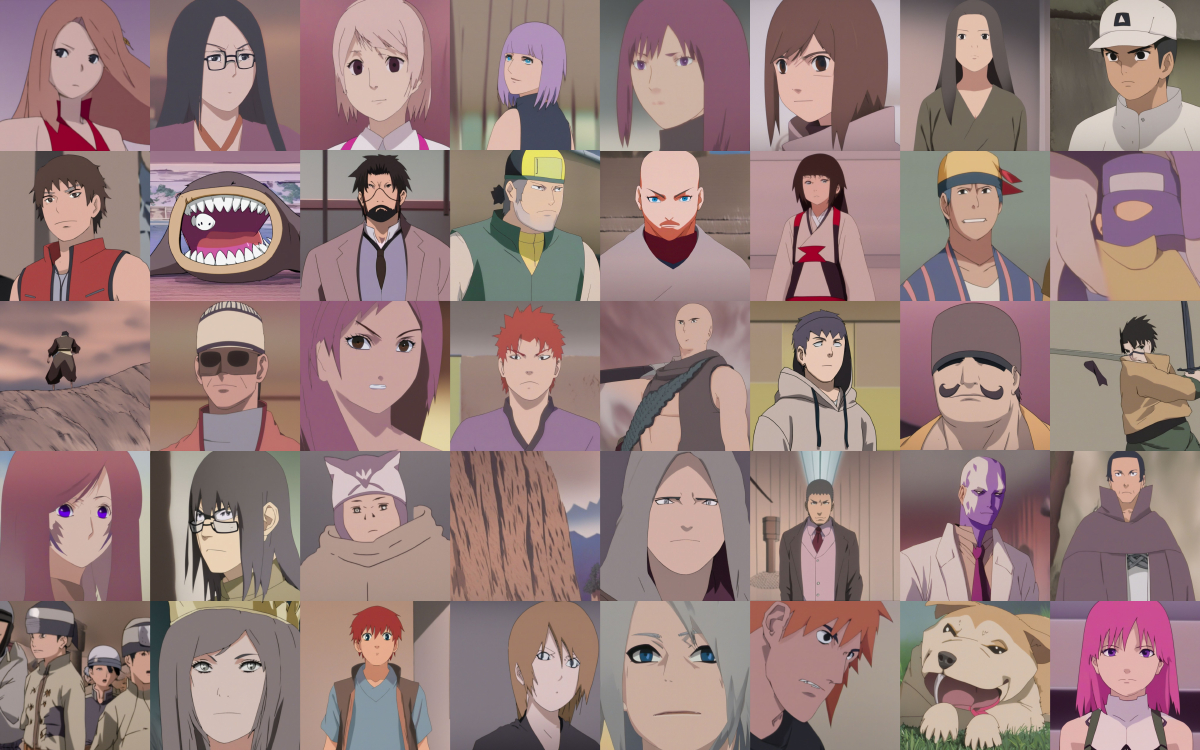}
    \caption{ OP-DoRA Naruto\citep{cervenka2022naruto2} results}
    \label{fig:naruto_opdora}
\end{figure*}

\begin{figure*}
    \centering
    \includegraphics[width=0.9\textwidth]{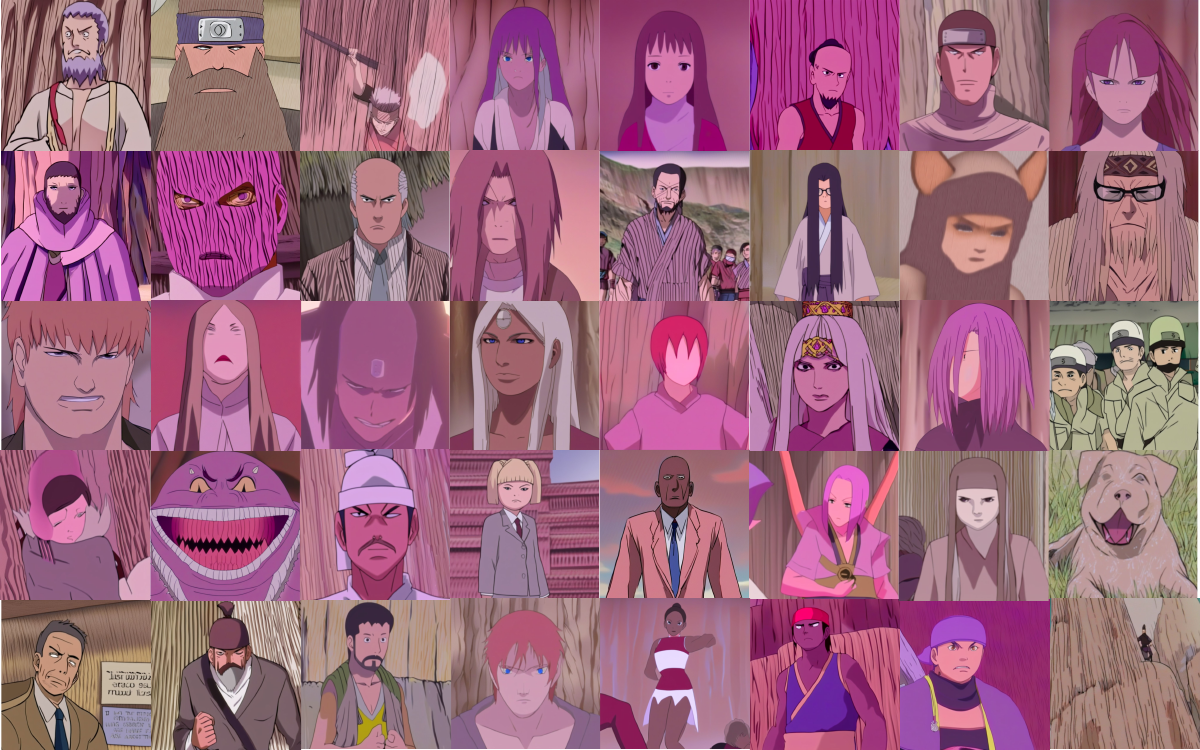}
    \caption{ DoRA Naruto\citep{cervenka2022naruto2} generated images}
    \label{fig:naruto_dora}
\end{figure*}

\begin{figure*}
    \centering
    \includegraphics[width=0.9\textwidth]{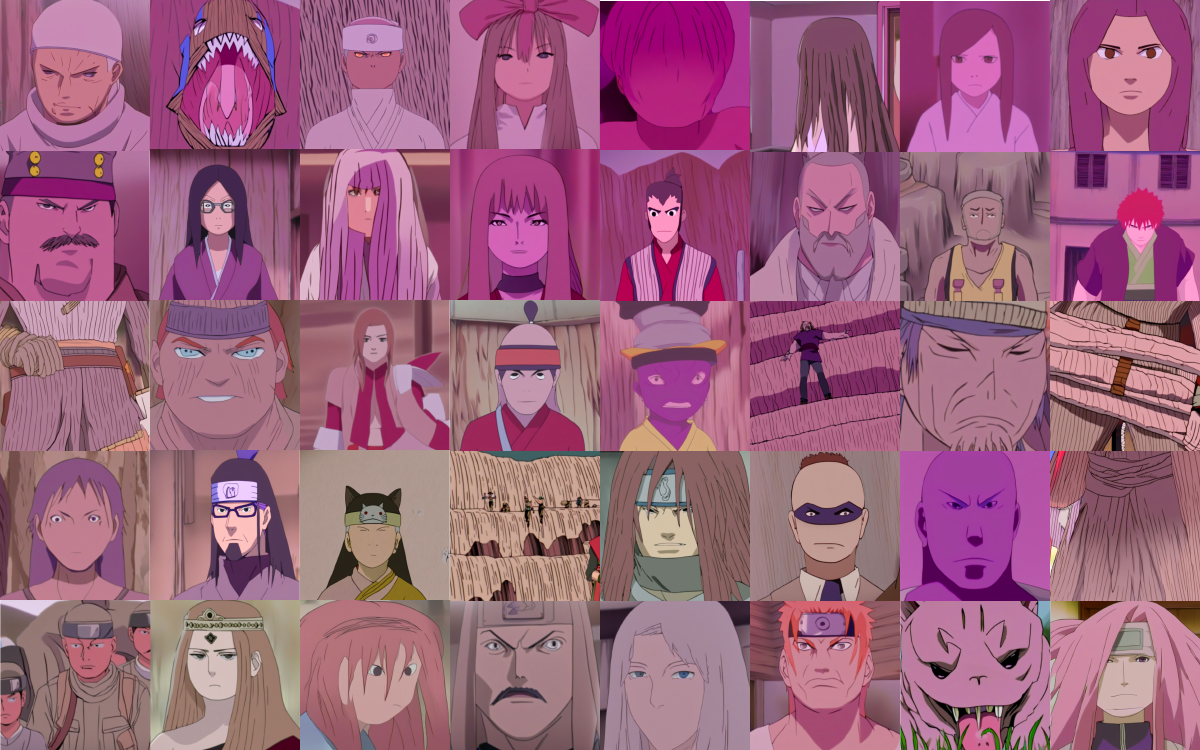}
    \caption{ LoRA Naruto\citep{cervenka2022naruto2} generated images}
    \label{fig:naruto_lora}
\end{figure*}

\begin{figure*}
    \centering
    \includegraphics[width=0.9\textwidth]{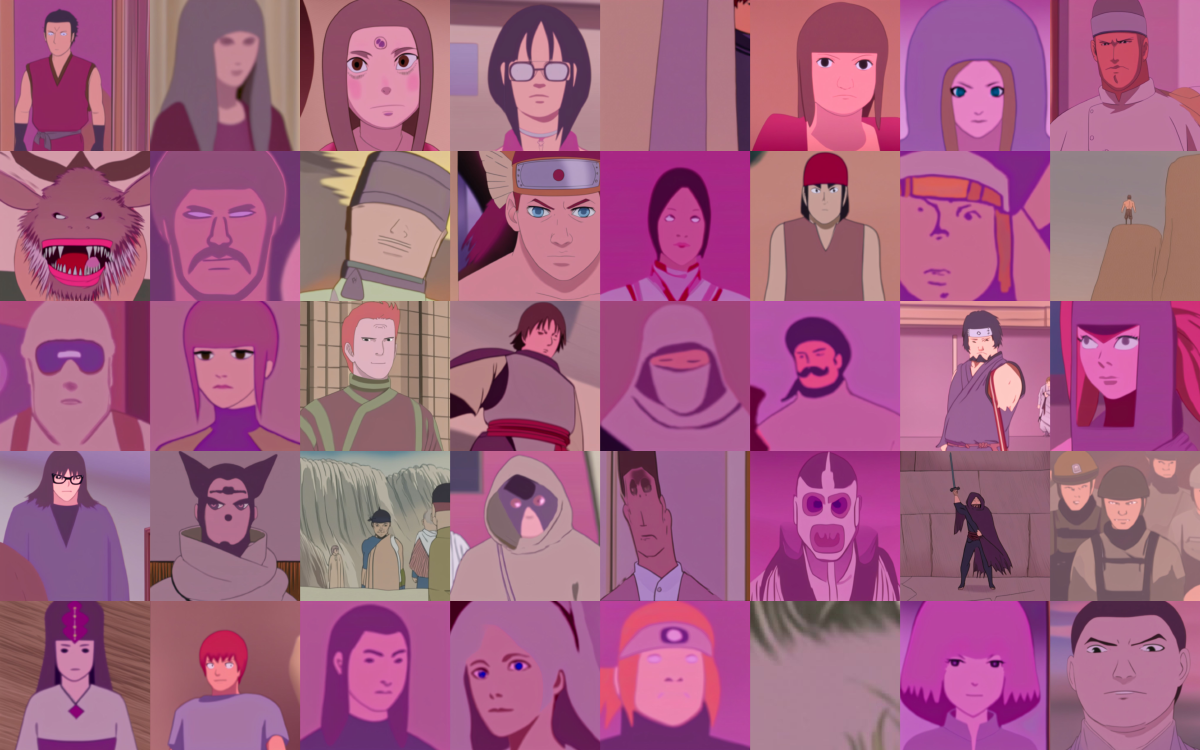}
    \caption{ PiSSA Naruto\citep{cervenka2022naruto2} generated images}
    \label{fig:naruto_pissa}
\end{figure*}

\begin{figure*}
    \centering
    \includegraphics[width=0.9\textwidth]{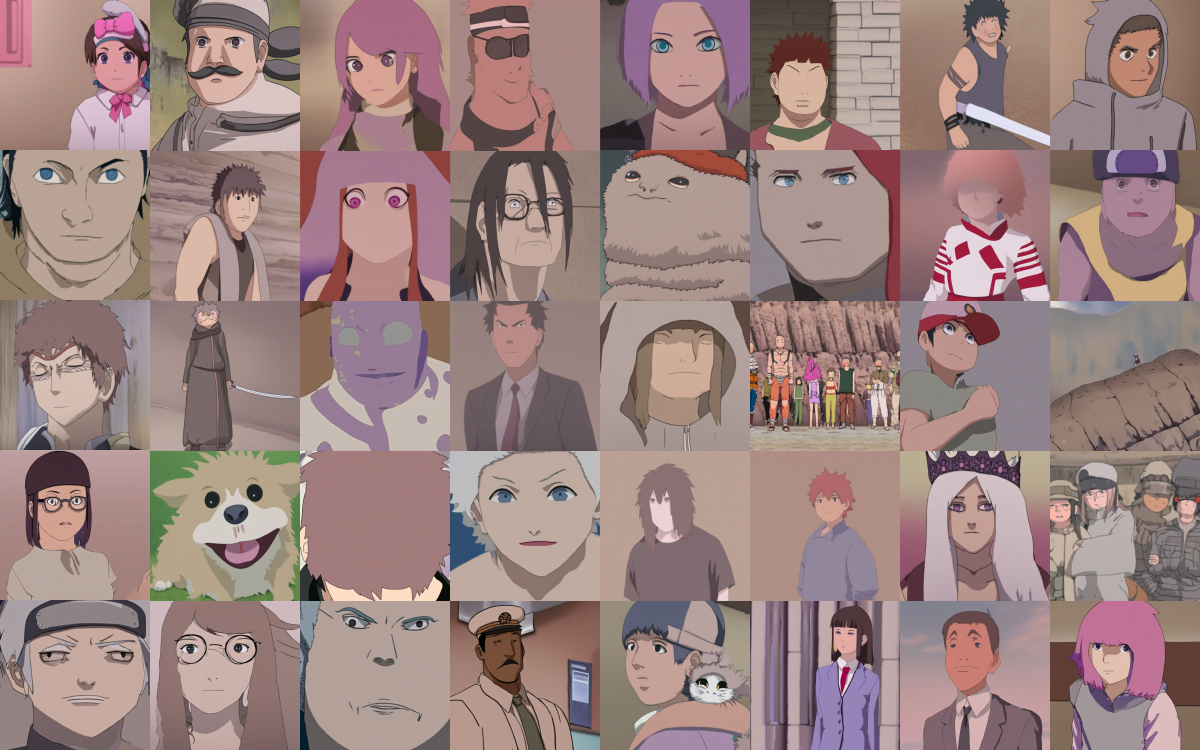}
    \caption{ LoRA-GA Naruto\citep{cervenka2022naruto2} generated images}
    \label{fig:naruto_ga}

\end{figure*}

\begin{figure*}
    \centering
    \includegraphics[width=0.9\textwidth]{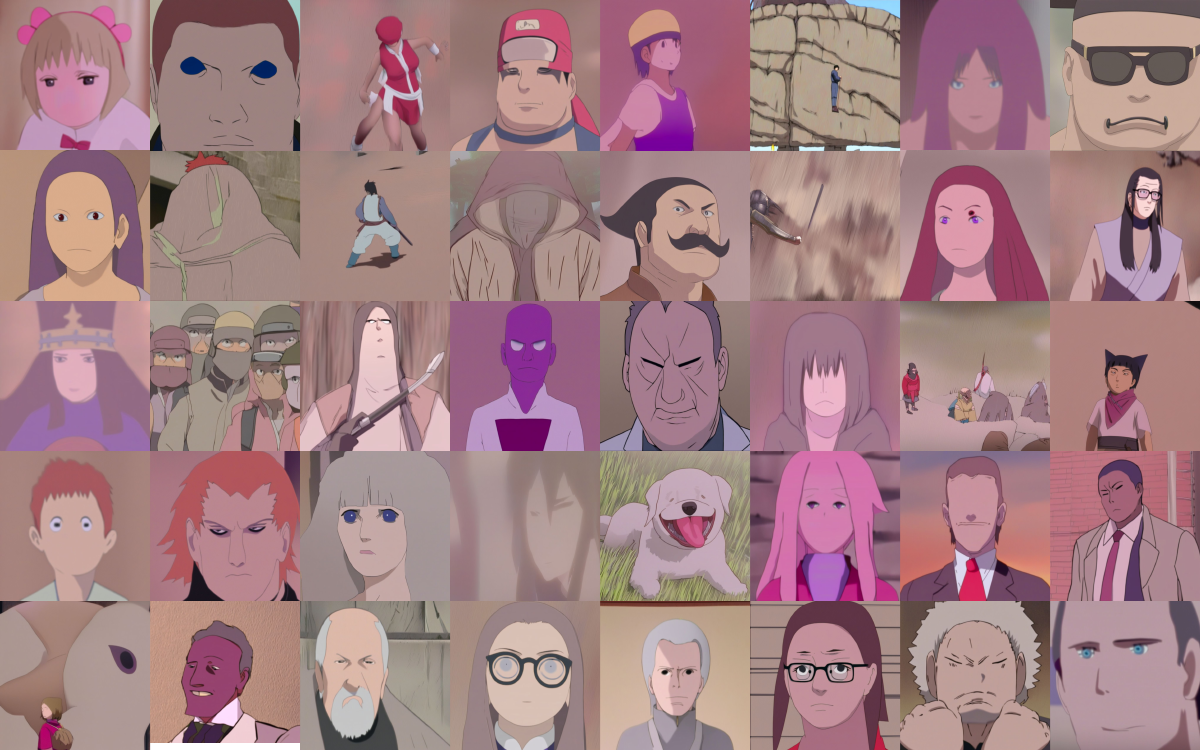}
    \caption{LoRA-PRO Naruto\citep{cervenka2022naruto2} generated images}
    \label{fig:naruto_pro}

\end{figure*}

\begin{figure*}
    \centering
    \includegraphics[width=0.9\textwidth]{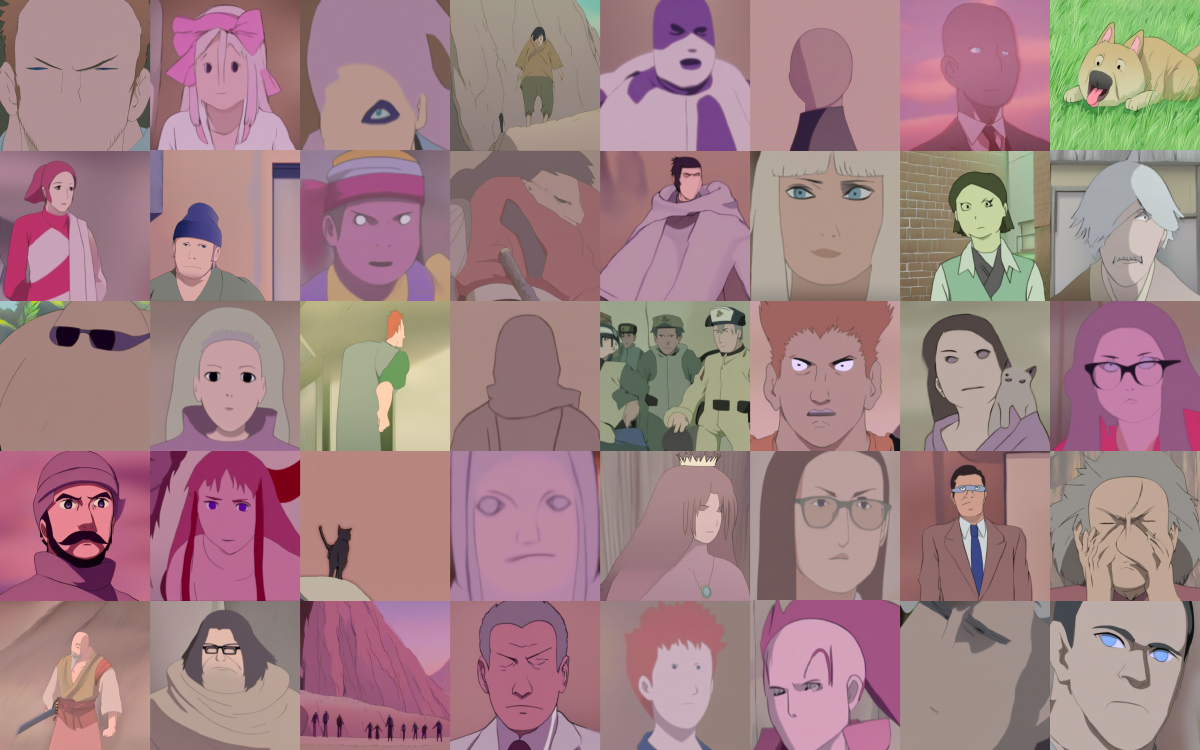}
    \caption{ScaledAdamW Naruto\citep{cervenka2022naruto2} generated images}
    \label{fig:naruto_scaledadamw}

\end{figure*}

\begin{figure*}
    \centering
    \includegraphics[width=0.9\textwidth]{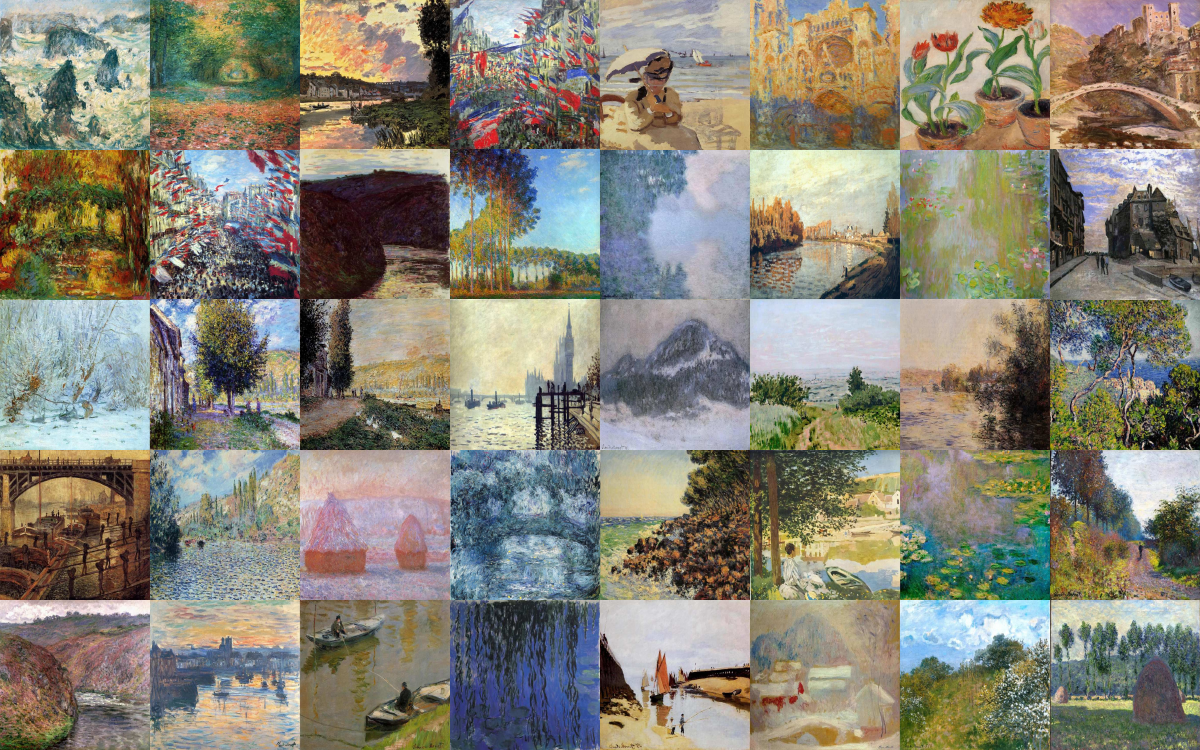}
    \caption{GT Monet WikiArt\citep{huggan2023wikiart} Images}
    \label{fig:monet_gt}
\end{figure*}

\begin{figure*}
    \centering
    \includegraphics[width=0.9\textwidth]{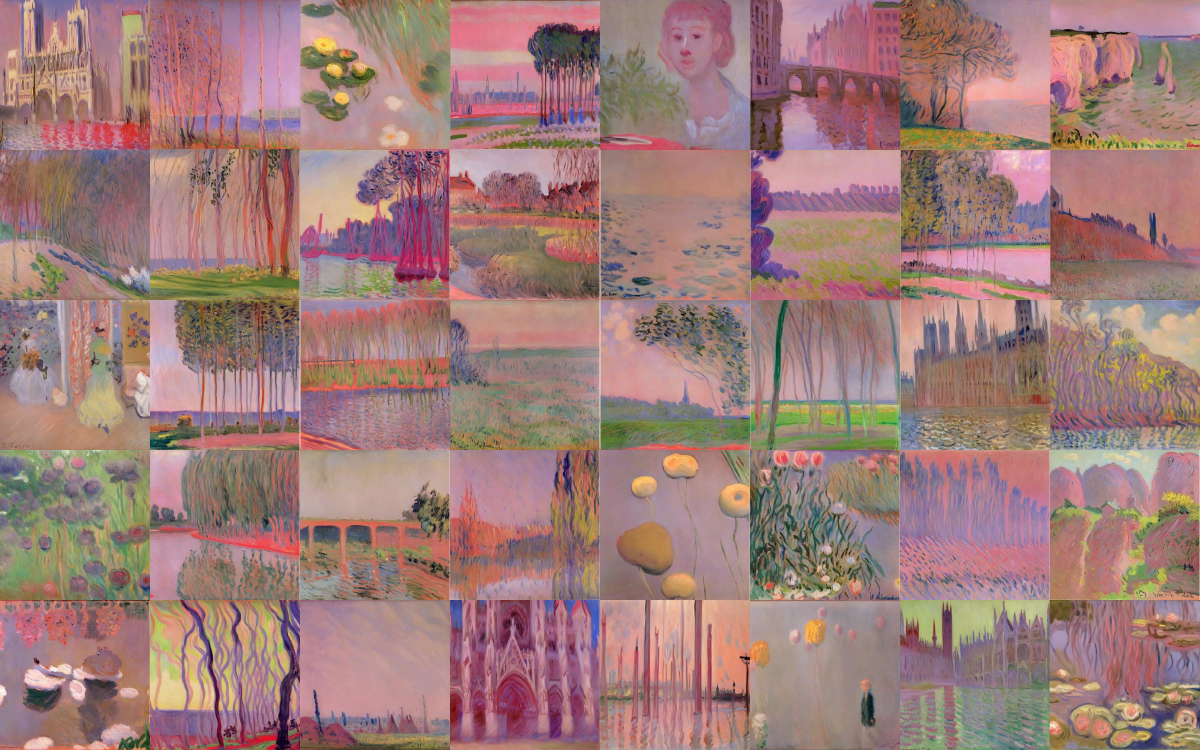}
    \caption{OP-LoRA Monet WikiArt\citep{huggan2023wikiart} Generated Images}
    \label{fig:monet_oplora}
\end{figure*}

\begin{figure*}
    \centering
    \includegraphics[width=0.9\textwidth]{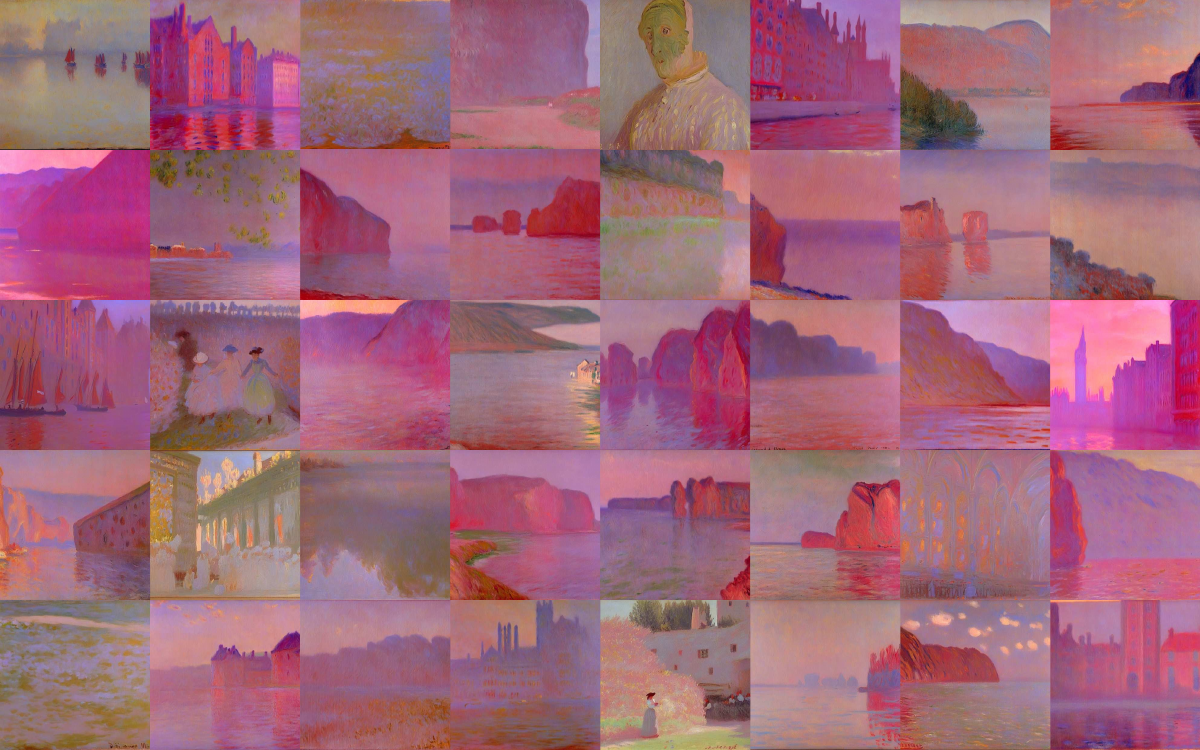}
    \caption{ OP-DoRA Monet WikiArt\citep{huggan2023wikiart} Generated Images}
    \label{fig:monet_opdora}

\end{figure*}

\begin{figure*}
    \centering
    \includegraphics[width=0.9\textwidth]{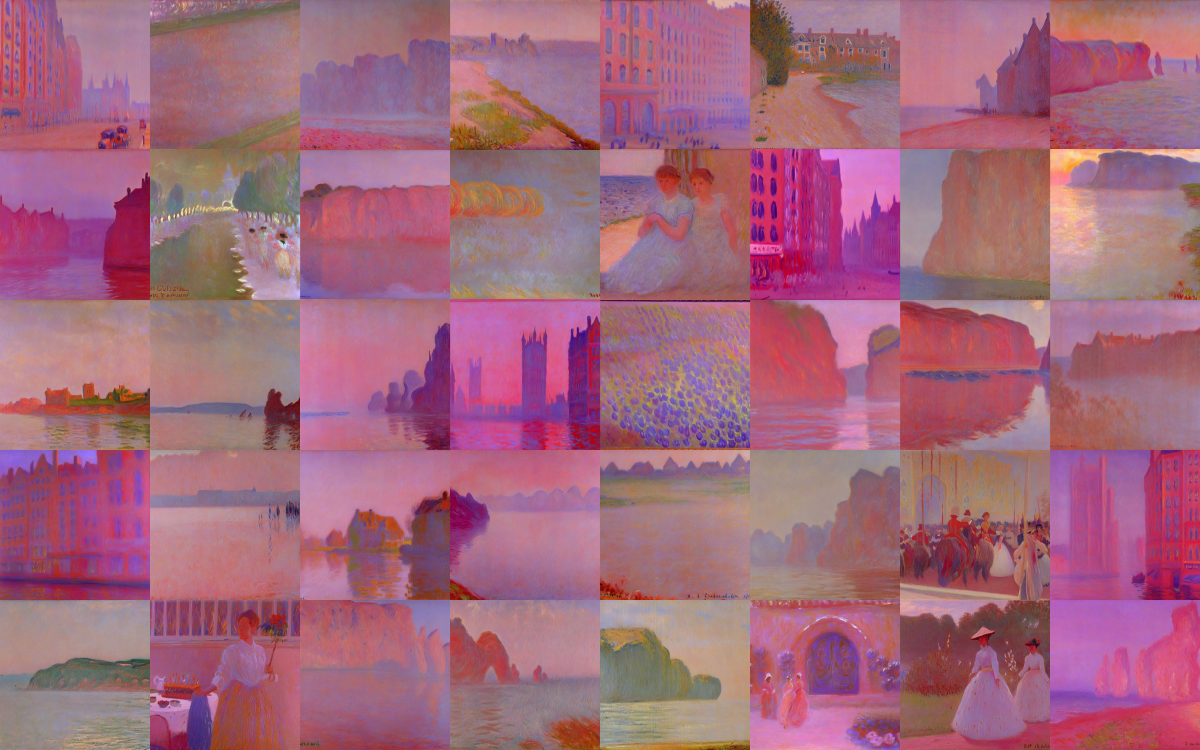}
    \caption{ DoRA Monet WikiArt\citep{huggan2023wikiart} Generated Images}
    \label{fig:monet_dora}
\end{figure*}

\begin{figure*}
    \centering
    \includegraphics[width=0.9\textwidth]{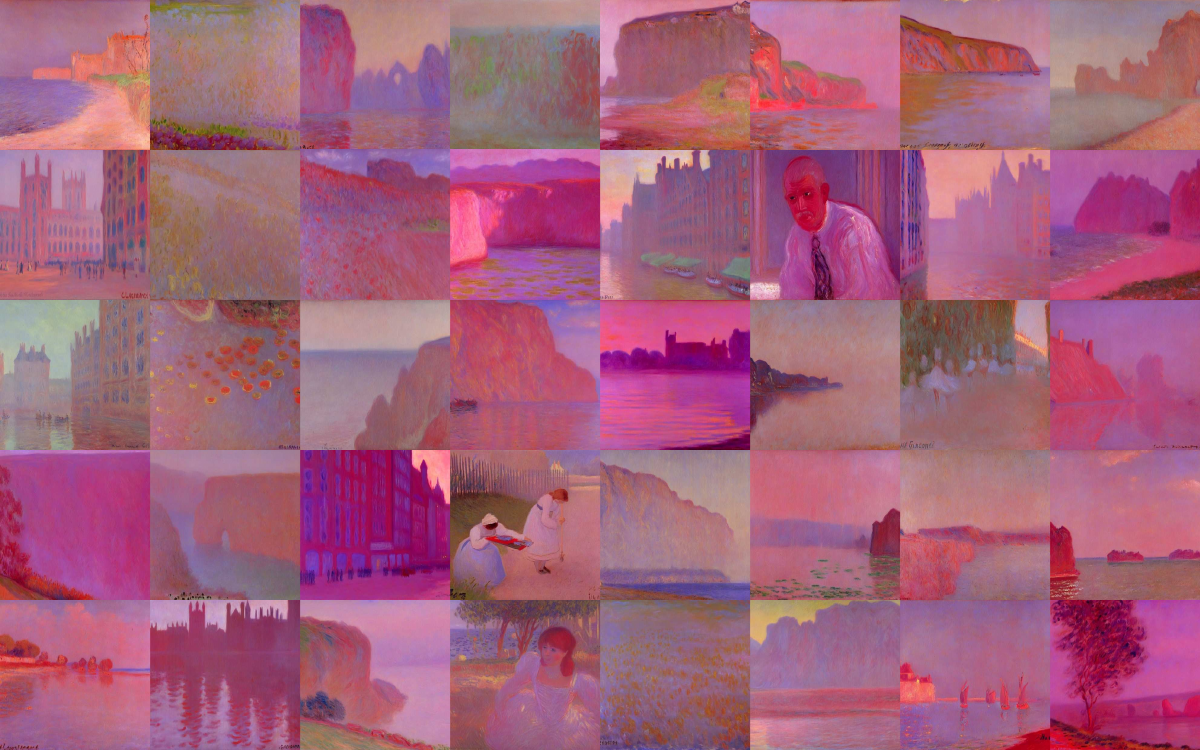}
    \caption{ LoRA Monet WikiArt\citep{huggan2023wikiart} Generated Images}
    \label{fig:monet_lora}
\end{figure*}

\begin{figure*}
    \centering
    \includegraphics[width=0.9\textwidth]{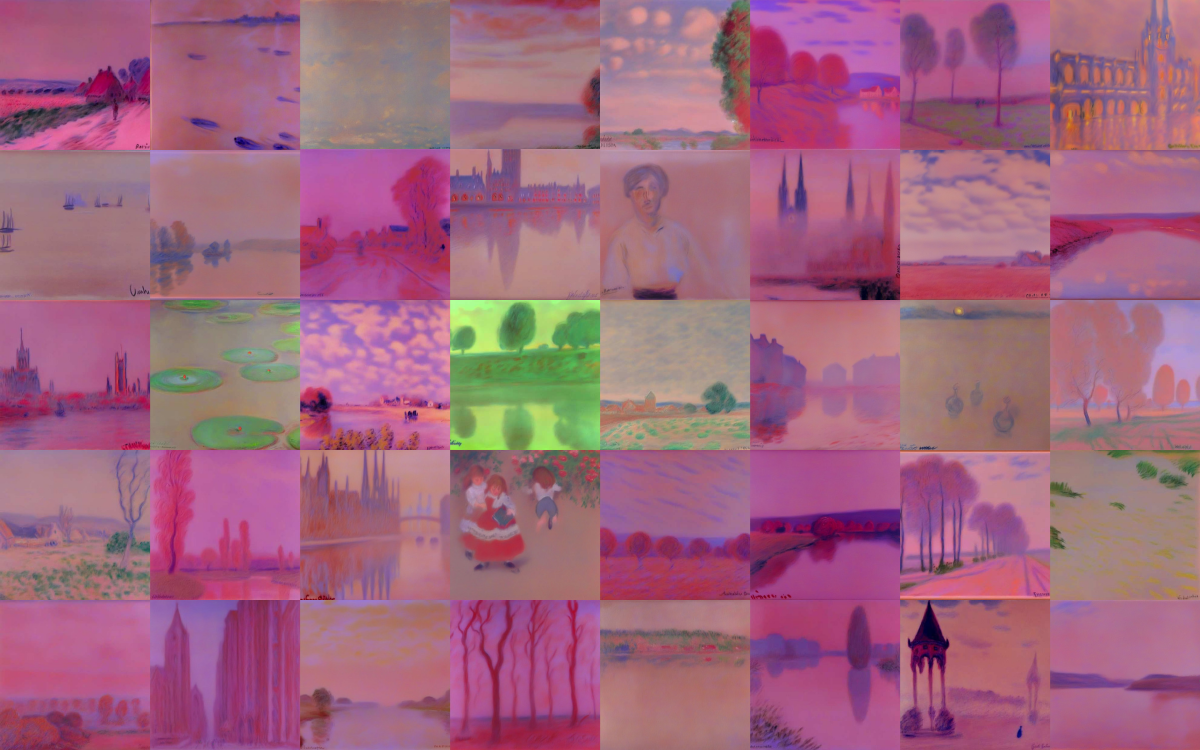}
    \caption{ PiSSA Monet WikiArt\citep{huggan2023wikiart} Generated Images}
    \label{fig:monet_pissa}

\end{figure*}

\begin{figure*}
    \centering
    \includegraphics[width=0.9\textwidth]{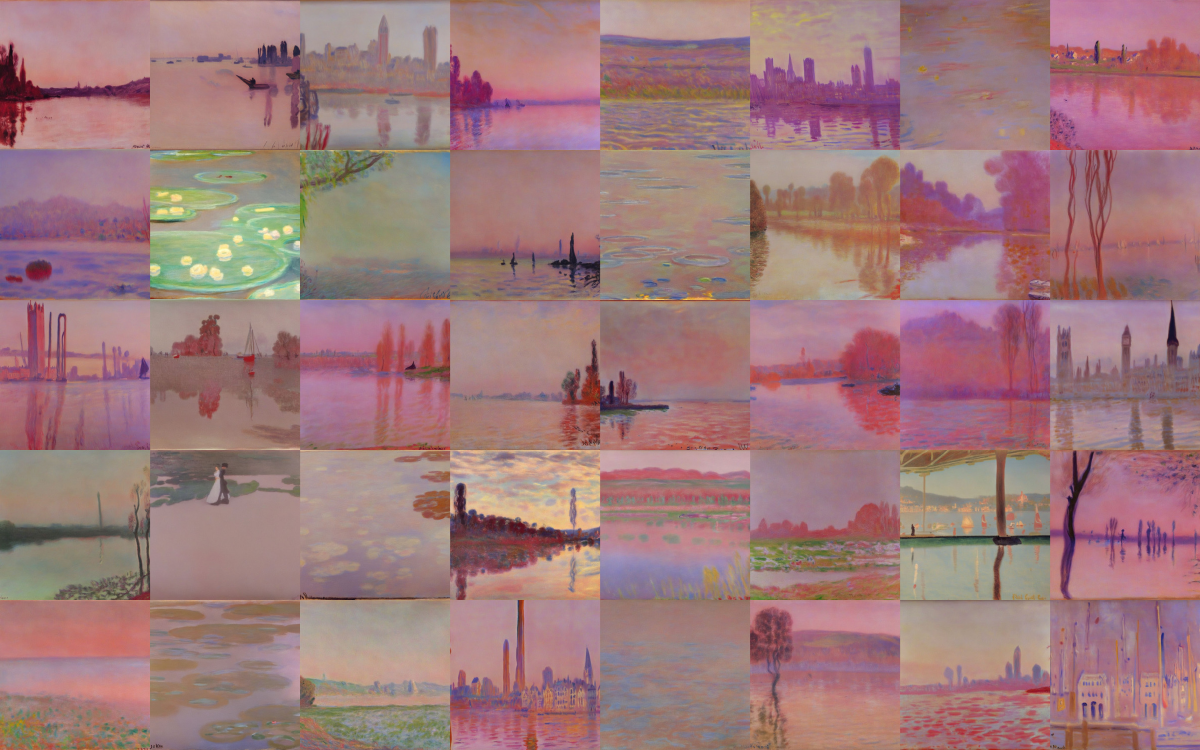}
    \caption{ LoRA-GA Monet WikiArt\citep{huggan2023wikiart} Generated Images}
    \label{fig:monet_ga}

\end{figure*}

\begin{figure*}
    \centering
    \includegraphics[width=0.9\textwidth]{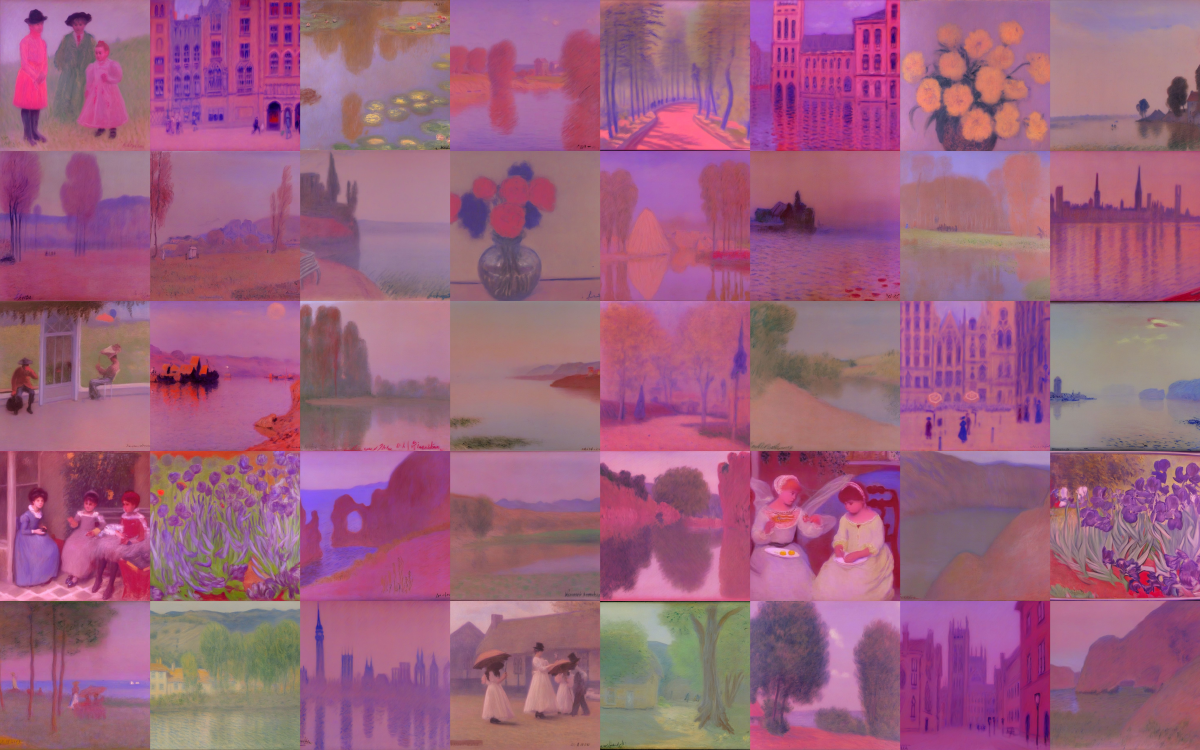}
    \caption{ LoRA-Pro Monet WikiArt\citep{huggan2023wikiart} Generated Images}
    \label{fig:monet_pro}
\end{figure*}

\begin{figure*}
    \centering
    \includegraphics[width=0.9\textwidth]{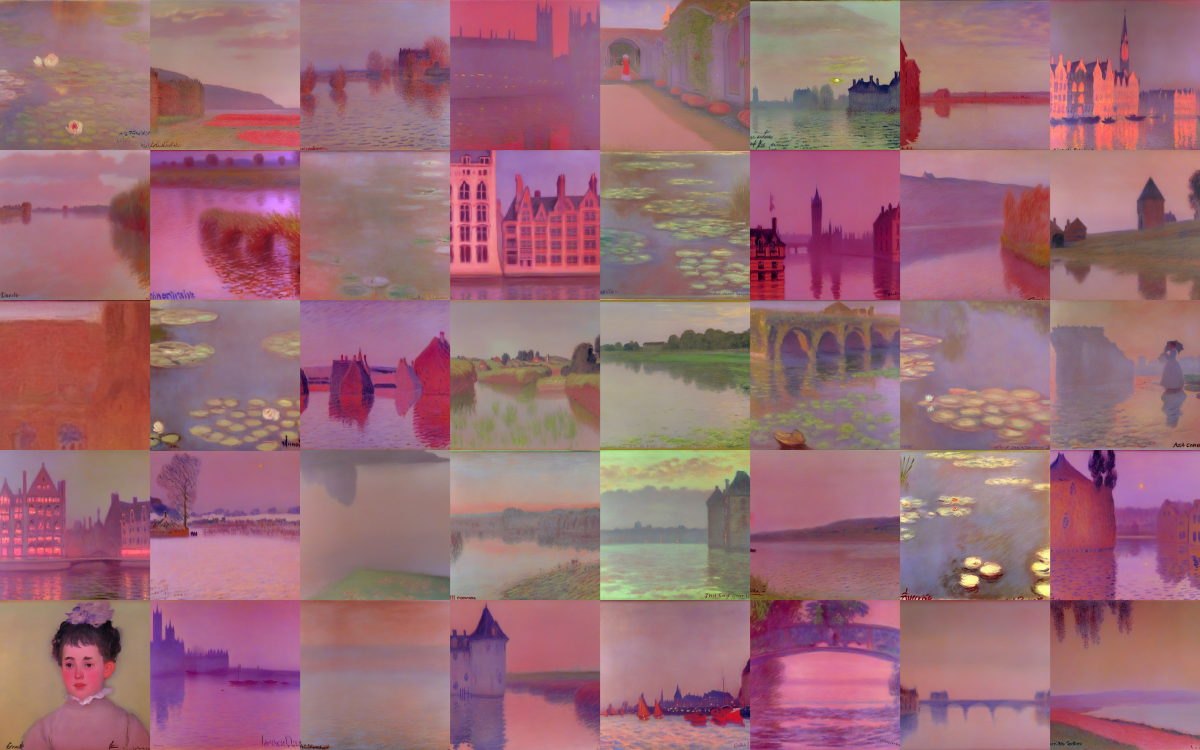}
    \caption{ ScaledAdamW Monet WikiArt\citep{huggan2023wikiart} Generated Images}
    \label{fig:monet_scaledadamw}
\end{figure*}

\section{Training Details}
\label{supp:training_details}

In this section, we summarize  the training settings of our main experiments.


\noindent\textbf{Hardware:} Most experiments were done with a single H100 80GB HB3 GPU.

\subsection{Initializing the OP-LoRA MLP}

In Section 3, we introduce the MLP used to predict low rank parameters as \[
\begin{aligned}
    \begin{pmatrix} A \\ B \end{pmatrix} &= W_2 (\text{ReLU}(W_1 z + c_1)) + c_2
\end{aligned}
\]

We initialize $W_1$ as Kaiming uniform. We also initalize $W_2$ as Kaiming uniform, except for parameters predicting the upsampling matrix B, which are initialized to zero to make the  initialization not change the pre-trained model behavior. All biases are initialized to 0. 

\subsection{Training Hyperparameters}

In Tables \ref{tab:sdxl_training_details} through \ref{tab:commonsense_training_details} we show training hyperparameters for experiments. These include batch sizes, learning rates, optimizer settings, and other configurations for  each task. For the main results in the paper, we adopt hyperparameters tuned for the LoRA baselines and apply the same settings to OP-LoRA, without any additional tuning specific to OP-LoRA. Nevertheless, OP-LoRA consistently attains strong performance under these inherited hyperparameters, indicating that it is robust and effective even without additional hyperparameter optimization.

\begin{table*}[ht!]
\centering
\begin{tabular}{lc}
\toprule
Hyperparameters  & All \\ \midrule
Base Model                  & Stable Diffusion XL 1.0\citep{podell2023sdxl}       \\ 

Rank \( r \)                    & 4       \\ 
\( \alpha \)                    & 4          \\ 
Dropout                         & 0.0         \\ 
Optimizer                       & AdamW        \\ 
LR                              & 1e-4        \\ 
LR Scheduler                    & Constant       \\ 
Batch size                      & 1           \\ 
Warmup Steps                    & 0          \\ 
Epochs                          & 2            \\ 
Where                           & U-Net Q, K, V, Out \\  
MLP-Width(OP-LoRA/OP-DoRA)                           & 32 \\  \bottomrule
\end{tabular}
\caption{Training Details for Stable Diffusion Finetuning Experiments.}
\label{tab:sdxl_training_details}
\end{table*}

\begin{table*}[ht!]
\centering
\begin{tabular}{lc}
\toprule
Hyperparameters  & All \\ \midrule
Base Model                  & VL-Bart\citep{cho2021unifying}       \\ 

Rank \( r \)                    & 128       \\ 
\( \alpha \)                    & 128          \\ 
Dropout                         & 0.0         \\ 
Optimizer                       & AdamW        \\ 
LR                              & 1e-3       \\ 
LR Scheduler                    & Linear       \\ 
Batch size                      & 300           \\ 
Warmup ratio                    & 0.1          \\ 
Epochs                          & 20            \\ 
Where                           & Q,K (Bias Also Trained) \\  
MLP-Width(OP-LoRA/OP-DoRA)                           & 32(OP-LoRA)/4(OP-DoRA) \\  \bottomrule
\end{tabular}
\caption{Training Details for VL-Bart Finetuning Experiments.}
\label{tab:vl_training_details}
\end{table*}

\begin{table*}[ht!]
\centering
\begin{tabular}{lc}
\toprule
Hyperparameters  & All \\ \midrule
Base Model                  & LLaVA1.5-7B\citep{liu2023llava}       \\ 

Rank \( r \)                    & 64/16       \\ 
\( \alpha \)                    & 128/32         \\ 
Dropout                         & 0.0         \\ 
Optimizer                       & AdamW        \\ 
LR                              & 5e-5        \\ 
LR Scheduler                    & Cosine       \\ 
Batch size                      & 64           \\ 
Warmup Ratio                    & 0 .03        \\ 
Epochs                          & 50            \\ 
Where                           & Multimodal Projector \\  
MLP-Width(OP-LoRA/OP-DoRA)                           & 768 \\  \bottomrule
\end{tabular}
\caption{Training Details for LLaVA Classification Finetuning Experiments.}
\label{tab:classification_training_details}
\end{table*}

\begin{table*}[ht!]
\centering
\begin{tabular}{lc}
\toprule
Hyperparameters  & All \\ \midrule
Base Model                  & LLaMA-7B\citep{touvron2023llama}       \\ 

Rank \( r \)                    & 32       \\ 
\( \alpha \)                    & 64          \\ 
Dropout                         & 0.05         \\ 
Optimizer                       & AdamW        \\ 
LR                              & 1e-4      \\ 
LR Scheduler                    & Linear       \\ 
Batch size                      & 16           \\ 
Warmup ratio                    & 0.03          \\ 
Epochs                          & 3            \\ 
Where                           & Q,K, V, Up, Down \\  
MLP-Width(OP-LoRA/OP-DoRA)                           & 32 \\  \bottomrule
\end{tabular}
\caption{Training Details for CommonSense Finetuning Experiments.}
\label{tab:commonsense_training_details}
\end{table*}